\documentclass[journal]{IEEEtran}
\usepackage{cite}

\ifCLASSINFOpdf
\else
\fi

\usepackage{amsmath,amssymb,bm,graphicx,color}
\allowdisplaybreaks[4]
\usepackage{url}

\begin{document}
%
\title{Large-scale Heteroscedastic Regression via Gaussian Process}
%
%
%

\author{Haitao~Liu,
	Yew-Soon~Ong,~\IEEEmembership{Fellow,~IEEE,}
	and~Jianfei~Cai,~\IEEEmembership{Senior~Member,~IEEE}
\thanks{Haitao Liu is with the Rolls-Royce@NTU Corporate Lab, Nanyang Technological University, Singapore, 637460. E-mail: htliu@ntu.edu.sg.}
\thanks{Yew-Soon Ong is with School of Computer Science and Engineering, Nanyang Technological University, Singapore, 639798. E-mail: asysong@ntu.edu.sg.

Jianfei Cai is with Department of Data Science \& AI, Monash University, Australia. E-mail: jianfei.cai@monash.edu.}
\thanks{}}

%
%

\markboth{}%
{Shell \MakeLowercase{\textit{et al.}}: Bare Demo of IEEEtran.cls for IEEE Journals}
%



\maketitle

\begin{abstract}
Heteroscedastic regression considering the varying noises among observations has many applications in the fields like machine learning and statistics. Here we focus on the heteroscedastic Gaussian process (HGP) regression which integrates the latent function and the noise function together in a unified non-parametric Bayesian framework. Though showing remarkable performance, HGP suffers from the cubic time complexity, which strictly limits its application to big data. To improve the scalability, we first develop a variational sparse inference algorithm, named VSHGP, to handle large-scale datasets. Furthermore, two variants are developed to improve the scalability and capability of VSHGP. The first is stochastic VSHGP (SVSHGP) which derives a factorized evidence lower bound, thus enhancing efficient stochastic variational inference. The second is distributed VSHGP (DVSHGP) which (i) follows the Bayesian committee machine formalism to distribute computations over multiple local VSHGP experts with many inducing points; and (ii) adopts hybrid parameters for experts to guard against over-fitting and capture local variety. The superiority of DVSHGP and SVSHGP as compared to existing scalable heteroscedastic/homoscedastic GPs is then extensively verified on various datasets.
\end{abstract}

\begin{IEEEkeywords}
Heteroscedastic GP, Large-scale, Sparse approximation, Stochastic variational inference, Distributed learning.
\end{IEEEkeywords}

%
\IEEEpeerreviewmaketitle

\section*{Notation}
\begin{align*}
F, F_V =& \mbox{Lower bounds for log model evidence} \log p(\bm{y}) \\
\bm{f}, \bm{g} =& \mbox{Latent function values and log variances} \\
\bm{f}_m, \bm{g}_u =& \mbox{Inducing variables for } \bm{f} \mbox{ and } \bm{g}\\
k^{f}, k^{g} =& \mbox{Kernels for } f \mbox{ and } g\\
\bm{K}^{f}, \bm{K}^{g} =& \mbox{Kernel matrices for } \bm{f} \mbox{ and } \bm{g} \\
m, u =& \mbox{Inducing sizes for } \bm{f} \mbox{ and } \bm{g} \\
m_0, u_0 =& \mbox{Inducing sizes for each expert in DVSHGP}\\
m_{\mathrm{b}} =& \mbox{Number of basis functions for GPVC [8]} \\
m_{\mathrm{sod}} =& \mbox{Subset size for EHSoD [13]} \\
M =& \mbox{Number of VSHGP experts} \\
\mathcal{M}_i =& i\mbox{th VSHGP expert } (1\le i \le M) \\
n, n_* =& \mbox{Training size and test size} \\
n_0 =& \mbox{Training size for each expert in DVSHGP} \\
\bm{X}, \bm{X}_* =& \mbox{Training and test data points} \\
\bm{y}, \bm{y}_* =& \mbox{Training and test observations} \\
\mu_0 =& \mbox{Mean parameter for latent noise function } g \\
\bm{\mu}_m, \bm{\Sigma}_m =& \mbox{Mean and variance for the posterior } q(\bm{f}_m)\\
\bm{\mu}_u, \bm{\Sigma}_u =& \mbox{Mean and variance for the posterior } q(\bm{g}_u)\\
\bm{\mu}_f, \bm{\Sigma}_f =& \mbox{Mean and variance for the posterior } q(\bm{f})\\
\bm{\mu}_f^{\star}, \bm{\Sigma}_f^{\star} =& \mbox{Mean and variance for the optimal } q^{\star}(\bm{f})\\
\bm{\mu}_g, \bm{\Sigma}_g =& \mbox{Mean and variance for the posterior } q(\bm{g})\\
\bm{\Lambda}_{n n} =& \mbox{Variational parameters for } q(\bm{g}_u) \mbox{ in VSHGP}\\
\mu_{f_{(\mathcal{A})}*}, \mu_{g_{(\mathcal{A})}*} =& \mbox{(Aggregated) prediction mean of } f \mbox{ and } g \mbox{ at } \bm{x}_*\\
\sigma^2_{f_{(\mathcal{A})}*}, \sigma^2_{g_{(\mathcal{A})}*} =& \mbox{(Aggregated) prediction variance of } f  \mbox{ and } g \mbox{ at } \bm{x}_*\\
\mu_{(\mathcal{A})*}, \sigma^2_{(\mathcal{A})*} =& \mbox{(Aggregated) prediction mean and variance at } \bm{x}_*\\
\mu_{i*}, \sigma^2_{i*} =& \mbox{Prediction mean and variance of } \mathcal{M}_i \mbox{ at } \bm{x}_*\\
w_{i*}^f, w_{i*}^g =& \mbox{Weights of the expert } \mathcal{M}_i \mbox{ at } \bm{x}_* \mbox{ for } f \mbox{ and } g \\
\bm{\theta}^f, \bm{\theta}^g =& \mbox{Kernel parameters for } k^f \mbox{ and } k^g \\
\epsilon_i =& \mbox{Noise for observation } y_i \mbox{ at point } \bm{x}_i \mbox{ } (1 \le i \le n)\\
\gamma =& \mbox{Step parameter for optimization}\\
\mathcal{B} =& \mbox{Mini-batch set sampled from } \bm{X}
\end{align*}

\section{Introduction}
%
%
%
%

\IEEEPARstart{I}{n} supervised learning, we learn a machine learning model from $n$ training points $\bm{X} = \{\bm{x}_i \in R^d\}_{i=1}^n$ and the observations $\bm{y} = \{y(\bm{x}_i) = f(\bm{x}_i) + \epsilon_i \}_{i=1}^n$, where $f$ is the underlying function and $\epsilon_i$ is the independent noise. The non-parametric Gaussian process (GP) places a GP prior on $f$ with the iid noise $ \epsilon_i \sim \mathcal{N}(0, \sigma^2_{\epsilon})$, resulting in the standard \textit{homoscedastic} GP. The homoscedasticity offers tractable GP inference to enable extensive applications in regression and classification \cite{rasmussen2006gaussian}, visualization \cite{lawrence2005probabilistic}, Bayesian optimization \cite{shahriari2016taking}, multi-task learning \cite{liu2018remarks}, active learning \cite{settles2012active}, etc.

Many realistic problems, e.g., volatility forecasting \cite{kou2015probabilistic}, biophysical variables estimation \cite{lazaro2014retrieval}, cosmological redshifts estimation \cite{almosallam2016gpz}, robotics and vehicle control \cite{bauza2017probabilistic, smith2018heteroscedastic}, however should consider the \textit{input-dependent} noise rather than the simple \textit{constant} noise in order to fit the local noise rates of complicated data distributions. In comparison to the conventional \textit{homogeneous} GP, the \textit{heteroscedastic} GP (HGP) provides better quantification of different sources of uncertainty, which further brings benefits for the downstream tasks, e.g., active learning, optimization and uncertainty quantification~\cite{kirschner2018information,kendall2017uncertainties}.

To account for the heteroscedastic noise in GP, there exists two main strategies: (i) treat the GP as a black-box and interpret the heteroscedasticity using another separate model; and (ii) integrate the heteroscedasticity within the unifying GP framework. The first post-model strategy first trains a standard GP to capture the underlying function $f$, and then either trains another GP to take the remaining empirical variance into account~\cite{urban2015sensor}, or use the quantile regression~\cite{pereira2014metamodel} to model the lower and upper quantiles of the variance respectively. 

In contrast, the second integration strategy provides an elegant framework for heteroscedastic regression. The simplest way to mimic variable noise is through adding independent yet different noise variances to the diagonal of kernel matrix~\cite{snelson2006variable}. Goldberg et al.~\cite{goldberg1998regression} introduced a more principled HGP which infers a mean-field GP for $f(\bm{x})$ and an additional GP $g(\bm{x})$ for $\log \sigma^2_{\epsilon}(\bm{x})$ jointly. Similarly, other HGPs which describe the heteroscedastic noise using the pointwise division of two GPs or the general non-linear combination of GPs have been proposed for example in~\cite{munoz2014divisive, munoz2016laplace,saul2016chained}. Note that unlike the homoscedastic GP, the inference in HGP is challenging since the model evidence (marginal likelihood) $p(\bm{y})$ and the posterior $p(y_* | \bm{X}, \bm{y}, \bm{x}_*)$ are intractable. Hence, various approximate inference methods, e.g., markov chain monte carlo (MCMC) \cite{goldberg1998regression}, maximum a posteriori (MAP) \cite{kersting2007most, quadrianto2009kernel, heinonen2016non, binois2018practical}, variational inference \cite{titsias2011variational, menictas2015variational}, expectation propagation~\cite{munoz2011heteroscedastic, tolvanen2014expectation} and Laplace approximation~\cite{hartmann2019laplace}, have been used. The most accurate MCMC is time-consuming when handling large-scale datasets; the MAP is a point estimation which risks over-fitting and oscillation; the variational inference and its variants, which run fast via maximizing over a tractable and rigorous evidence lower bound (ELBO), provide a trade-off. 

This paper focuses on the HGP model developed in~\cite{goldberg1998regression}. When handling $n$ training points, the standard GP suffers from a cubic complexity $\mathcal{O}(n^3)$ due to the inversion of an $n \times n$ kernel matrix, which makes it unaffordable for big data. Since HGP employs an additional log-GP for noise variance, its complexity is about two times that of standard GP. Hence, to handle large-scale datasets, which is of great demand in the era of big data, the scalability of HGP should be improved.

Recently, there has been an increasing trend on studying scalable GPs, which have two core categories: global and local approximations~\cite{liu2020when}. As the representative of global approximation, sparse approximation considers $m$ ($m \ll n$) \textit{global} inducing pairs $\{\bm{X}_m, \bm{f}_m\}$ to summarize the training data by approximating the prior \cite{quinonero2005unifying} or the posterior \cite{titsias2009variational}, resulting in the complexity of $\mathcal{O}(nm^2)$. Variants of sparse approximation have been recently proposed to handle millions of data points via distributed inference \cite{gal2014distributed, hoang2016distributed}, stochastic variational inference \cite{hensman2013gaussian, hoang2015unifying}, or structured inducing points \cite{wilson2015kernel}. The global sparse approximation, however, often faces challenges in capturing quick-varying features~\cite{bui2014tree}. Differently, local approximation, which follows the idea of divide-and-conquer, first trains GP experts on local subsets and then aggregates their predictions, for example by the means of product-of-experts (PoE) \cite{hinton2002training} and Bayesian committee machine (BCM) \cite{tresp2000bayesian, deisenroth2015distributed, liu2018generalized}. Hence, local approximation not only distributes the computations but also captures quick-varying features~\cite{liu2019understanding}. Hybrid strategies thereafter have been presented for taking advantages of global and local approximations~\cite{snelson2007local, vanhatalo2008modelling, ouyang2018gaussian}.

The developments in scalable homoscedastic GPs have thus motivated us to scale up HGPs. Alternatively, we could combine the simple Subset-of-Data (SoD) approximation~\cite{chalupka2013framework} with the empirical HGP~\cite{urban2015sensor} which trains two separate GPs for predicting mean and variance, respectively. The empirical variance however is hard to fit since it follows an asymmetric Gaussian distribution. More reasonably, the GP using variable covariances (GPVC)~\cite{almosallam2016gpz} follows the idea of relevance vector machine (RVM) \cite{tipping2001sparse} that a stationary kernel $k(.,.)$ has a positive and finite Fourier spectrum, suggesting using only $m_{\mathrm{b}}$ ($m_{\mathrm{b}} \ll n$) independent basis functions for approximation. Note that GPVC shares the basis functions for $f$ and $g$ which however might produce distinct features. Besides, the RVM-type model usually suffers from underestimated prediction variance when leaving $\bm{X}$ \cite{rasmussen2005healing}.

Besides, there are some scalable ``pseudo''  HGPs~\cite{snelson2006sparse, snelson2007local} which are not designed for such case, but can describe the heteroscedasticity to some extent due to the factorized conditionals. For instance, the Fully Independent Training Conditional (FITC)~\cite{snelson2006sparse} and its block version named Partially Independent Conditional (PIC)~\cite{snelson2007local} adopt a factorized training conditional to achieve a varying noise~\cite{bauer2016understanding}. Though equipped with heteroscedastic variance, these models (i) severely underestimate the constant noise variance, (ii) sacrifice the prediction mean~\cite{bauer2016understanding}, and (iii) may produce discontinuous predictions on block boundaries~\cite{vanhatalo2008modelling}. Recently, the stochastic/distributed variants of FITC and PIC have been developed to further improve the scalability~\cite{hoang2015unifying, hoang2016distributed, yu2019stochastic, ouyang2018gaussian}. 

The high capability of describing complicated data distribution with however poor scalability motivate us to develop variational sparse HGPs which employ an additional log-GP for heteroscedasticity. Particularly, the main contributions are:
 
\textit{1}. A variational inference algorithm for sparse HGP, named VSHGP, is developed. Specifically, VSHGP derives an analytical ELBO using $m$ inducing points for $f$ and $u$ inducing points for $g$, resulting in a greatly reduced complexity of $\mathcal{O}(nm^2 + nu^2)$. Besides, some tricks for example re-parameterization are used to ease the inference;

\textit{2}. The stochastic variant SVSHGP is presented to further improve the scalability of VSHGP. Specifically, we derive a factorized ELBO which allows using efficient stochastic variational inference;

\textit{3}. The distributed variant DVSHGP is presented for improving the scalability and capability of VSHGP. The local experts with many inducing points (i) distribute the computations for parallel computing, and (ii) employ hybrid parameters to guard against over-fitting and capture local variety;

\textit{4}. Extensive experiments\footnote{The SVSHGP is implemented based on the GPflow toolbox~\cite{matthews2017gpflow}, which benefits from parallel/GPU speedup and automatic differentiation of Tensorflow~\cite{abadi2016tensorflow}. The DVSHGP is built upon the GPML toolbox~\cite{rasmussen2010gaussian}. These implementations are available at \url{https://github.com/LiuHaiTao01}.} conducted on datasets with up to two million points reveal that the localized DVSHGP exhibits superior performance, while the global SVSHGP may sacrifice the prediction mean for capturing heteroscedastic noise. 

The remainder of the article is organized as follows. Section~\ref{sec_VSHGP} first develops the VSHGP model via variational inference. Then, we present the stochastic variant in Section~\ref{sec_SVSHGP} and the distributed variant in Section~\ref{sec_DVSHGP} to further enhance the scalability and capability, followed by extensive experiments in Section~\ref{sec_results}. Finally, Section~\ref{sec_conclusion} provides concluding remarks.

\section{Variational sparse HGP} \label{sec_VSHGP}
\subsection{Sparse approximation}
We follow~\cite{goldberg1998regression} to define the HGP as $y(\bm{x})=f(\bm{x})+\epsilon(\bm{x})$, wherein the latent function $f(\bm{x})$ and the noise $\epsilon(\bm{x})$ follow
\begin{equation} \label{eq_HGP}
f(\bm{x}) \sim \mathcal{GP}(0, k^f(\bm{x}, \bm{x}')), \quad \epsilon(\bm{x}) \sim \mathcal{N}(0, \sigma^2_{\epsilon}(\bm{x})).
\end{equation}
It is observed that the input-dependent noise variance $\sigma^2_{\epsilon}(\bm{x})$ enables describing possible heteroscedasticity. Notably, this HGP degenerates to homoscedastic GP when $\sigma^2_{\epsilon}(\bm{x})$ is a constant. To ensure the positivity, we particularly consider the exponential form $\sigma^2_{\epsilon}(\bm{x}) = e^{g(\bm{x})}$, wherein the latent function $g(\bm{x})$ akin to $f(\bm{x})$ follows an independent GP prior
\begin{equation}
g(\bm{x}) \sim \mathcal{GP}(\mu_0, k^g(\bm{x}, \bm{x}')).
\end{equation}
The only difference is that unlike the zero-mean GP prior placed on $f(\bm{x})$, we explicitly consider a prior mean $\mu_0$ to account for the variability of the noise variance.\footnote{For $f$, we can pre-process the data to fulfill the zero-mean assumption. For $g$, however, it is hard to satisfy the zero-mean assumption, since we have no access to the ``noise'' data.} The kernels $k^f$ and $k^g$ could be, e.g., the squared exponential (SE) kernel equipped with automatic relevance determination (ARD)
\begin{equation} \label{eq_SE_Kernel}
k(\bm{x},\bm{x}')= \sigma^2_{s} \exp \left(- \frac{1}{2} \sum_{i=1}^d \frac{(x_i-x'_i)^2}{l_i^2} \right),
\end{equation}
where the signal variance $\sigma^2_{s}$ is an output scale, and the length-scale $l_i$ is an input scale along the $i$th dimension. 

Given the training data $\mathcal{D} = \{\bm{X}, \bm{y} \}$, the joint priors follow
\begin{equation}
p(\bm{f}) = \mathcal{N} (\bm{f} | \bm{0}, \bm{K}^f_{nn}), \quad p(\bm{g}) = \mathcal{N} (\bm{g} | \mu_0 \bm{1}, \bm{K}^g_{nn}), 
\end{equation}
where $[\bm{K}^f_{nn}]_{ij} = k^f(\bm{x}_i, \bm{x}_j)$ and $[\bm{K}^g_{nn}]_{ij} = k^g(\bm{x}_i, \bm{x}_j)$ for $1 \le i,j \le n$. Accordingly, the data likelihood becomes
\begin{equation}
p(\bm{y} | \bm{f}, \bm{g}) = \mathcal{N}(\bm{y} | \bm{f}, \bm{\Sigma}_{\epsilon}),
\end{equation}
where the diagonal noise matrix has $[\bm{\Sigma}_{\epsilon}]_{ii} = e^{g(\bm{x}_i)}$.

To scale up HGP, we follow the sparse approximation framework to introduce $m$ inducing variables $\bm{f}_m \sim \mathcal{N}(\bm{f}_m|\bm{0}, \bm{K}_{mm}^f)$ at the inducing points $\bm{X}_m$ for $\bm{f}$; similarly, we introduce $u$ inducing variables $\bm{g}_{u}  \sim \mathcal{N}(\bm{g}_{u}|\mu_0 \bm{1}, \bm{K}_{uu}^g)$ at the independent $\bm{X}_{u}$ for $\bm{g}$. Besides, we assume that $\bm{f}_m$ is a sufficient statistic for $\bm{f}$, and $\bm{g}_{u}$ a sufficient statistic for $\bm{g}$.\footnote{Sufficient statistic means the variables $\bm{z}$ and $\bm{f}$ are independent given $\bm{f}_m$, i.e., it holds $p(\bm{z}| \bm{f}, \bm{f}_m) = p(\bm{z}| \bm{f}_m)$.} As a result, we obtain two training conditionals 
\begin{align*}
	p(\bm{f}|\bm{f}_m) &= \mathcal{N} (\bm{f}|\bm{\Omega}_{nm}^f \bm{f}_m, \bm{K}_{nn}^f - \bm{Q}_{nn}^f), \\
	p(\bm{g}|\bm{g}_{u}) &= \mathcal{N} (\bm{g}|\bm{\Omega}_{nu}^g (\bm{g}_{u} -  \mu_0\bm{1}) + \mu_0\bm{1}, \bm{K}_{nn}^g - \bm{Q}_{nn}^g),
\end{align*}
where $\bm{\Omega}_{nm}^{f} = \bm{K}_{nm}^{f} (\bm{K}_{mm}^{f})^{-1}$, $\bm{\Omega}_{nm}^{g} = \bm{K}_{nm}^{g} (\bm{K}_{mm}^{g})^{-1}$, $\bm{Q}_{nn}^{f} = \bm{\Omega}_{nm}^{f} \bm{K}_{mn}^{f}$ and $\bm{Q}_{nn}^{g} = \bm{\Omega}_{nm}^{g} \bm{K}_{mn}^{g}$. 

In the augmented probability space, the model evidence
\begin{equation*}
p(\bm{y}) = \int p(\bm{y} | \bm{f}, \bm{g}) p(\bm{f}|\bm{f}_m) p(\bm{g}|\bm{g}_{u}) p(\bm{f}_m) p(\bm{g}_u) d\bm{f} d\bm{f}_m d\bm{g} d\bm{g}_u,
\end{equation*}
together with the posterior $p(\bm{z}|\bm{y}) = p(\bm{y}|\bm{z}) p(\bm{z}) / p(\bm{y})$ where $\bm{z} = \{\bm{f}, \bm{g}, \bm{f}_m, \bm{g}_{u}\}$, however, is intractable. Hence, we use variational inference to derive an analytical ELBO of $\log p(\bm{y})$.

\subsection{Evidence lower bound}
We employ the mean-field theory \cite{sun2013review} to approximate the intractable posterior $p(\bm{z}| \bm{y}) = p(\bm{f}|\bm{f}_m) p(\bm{g}| \bm{g}_{u}) p(\bm{f}_m| \bm{y}) p(\bm{g}_{u}| \bm{y})$ as
\begin{equation}
p(\bm{z}| \bm{y}) \approx q(\bm{z}) = p(\bm{f}|\bm{f}_m) p(\bm{g}| \bm{g}_{u}) q(\bm{f}_m) q(\bm{g}_{u}),
\end{equation}
where $q(\bm{f}_m)$ and $q(\bm{g}_{u})$ are free variational distributions to approximate the posteriors $p(\bm{f}_m | \bm{y})$ and $p(\bm{g}_{u} | \bm{y})$, respectively. 

In order to push the approximation $q(\bm{z})$ towards the exact $p(\bm{z}| \bm{y})$, we minimize their Kullback-Leibler (KL) divergence $\mathrm{KL}(q(\bm{z}) || p(\bm{z}| \bm{y}))$, which, on the other hand, is equivalent to maximizing the ELBO $F$, since $\mathrm{KL}(.,.) \geq 0$, as
\begin{equation} \label{eq_VB}
F = \int q(\bm{z}) \log \frac{p(\bm{z}, \bm{y})}{q(\bm{z})} d\bm{z} = \log p(\bm{y}) - \mathrm{KL}(q(\bm{z}) || p(\bm{z}| \bm{y})).
\end{equation}
As a consequence, instead of directly maximizing the intractable $\log p(\bm{y})$ for inference, we now seek the maximization of $F$ w.r.t. the variational distributions $q(\bm{f}_m)$ and $q(\bm{g}_{u})$.

By reformulating $F$ we observe that
\begin{align*}
F = -\mathrm{KL}(q(\bm{f}_m) || q^{\star}(\bm{f}_m)) + H(q(\bm{g}_{u})) + const.,
\end{align*}
where $H(q(\bm{g}_{u}))$ is the information entropy of $q(\bm{g}_u)$; $q^{\star}(\bm{f}_m)$ is the optimal distribution since it maximizes the bound $F$, and it satisfies, given the normalizer $C_0$,
\begin{equation} \label{e1_q*(fm)}
q^{\star}(\bm{f}_m) = \frac{p(\bm{f}_m)}{C_0} e^{\int p(\bm{f}|\bm{f}_m) p(\bm{g}| \bm{g}_{u}) q(\bm{g}_{u}) \log p(\bm{y} | \bm{f}, \bm{g}) d\bm{f} d\bm{g} d\bm{g}_{u}}.
\end{equation}

Thereafter, by substituting $q^{\star}(\bm{f}_m)$ back into $F$, we arrive at a tighter ELBO, given $q(\bm{g}_{u}) = \mathcal{N} (\bm{g}_{u}|\bm{\mu}_{u}, \bm{\Sigma}_{u})$, as
\begin{equation} \label{eq_F}
\begin{aligned}
F_V =&  \log C_0 - \mathrm{KL}(q(\bm{g}_{u}) || p(\bm{g}_{u})) \\
=& \underbrace{\log \mathcal{N}(\bm{y}|\bm{0}, \bm{Q}_{nn}^f + \bm{R}_g)}_{\mathrm{log \, term}} - \underbrace{0.25 \mathrm{Tr}[\bm{\Sigma}_{g}]}_{\mathrm{trace \, term \, of} \, \bm{g}} \\
&- \underbrace{0.5 \mathrm{Tr}[\bm{R}_g^{-1}(\bm{K}_{nn}^f - \bm{Q}_{nn}^f)]}_{\mathrm{trace \, term \, of} \, \bm{f}} - \underbrace{\mathrm{KL}(q(\bm{g}_{u}) || p(\bm{g}_{u}))}_{\mathrm{KL \, term}},
\end{aligned}
\end{equation}
where the diagonal matrix $\bm{R}_g \in R^{n \times n}$ has $[\bm{R}_g]_{ii} = e^{[\bm{\mu}_{g}]_i - [\bm{\Sigma}_{g}]_{ii}/2}$, with the mean and variance
\begin{subequations}
	\begin{align}
	\bm{\mu}_{g} &= \bm{\Omega}_{nu}^g (\bm{\mu}_{u} - \mu_0\bm{1}) + \mu_0\bm{1}, \\
	\bm{\Sigma}_{g} &= \bm{K}_{nn}^g - \bm{Q}_{nn}^g + \bm{\Omega}_{nu}^g \bm{\Sigma}_{u} (\bm{\Omega}_{nu}^g)^{\mathsf{T}},
	\end{align}
\end{subequations}
coming from $q(\bm{g}) = \int p(\bm{g}|\bm{g}_u) q(\bm{g}_u) d\bm{g}_u$ which approximates $p(\bm{g}|\bm{y})$. It is observed that the analytical bound $F_V$ depends only on $q(\bm{g}_{u})$ since we have ``marginalized'' $q(\bm{f}_m)$ out. 

Let us delve further into the terms of $F_V$ in~\eqref{eq_F}:
\begin{itemize}
	\item The log term $\log \mathcal{N}(\bm{y}|\bm{0}, \bm{\Sigma}_y)$, where $\bm{\Sigma}_y = \bm{Q}_{nn}^f + \bm{R}_g$, is analogous to that in a standard GP. It achieves bias-variance trade-off for both $f$ and $g$ by penalizing model complexity and low data likelihood \cite{rasmussen2006gaussian}.
	\item The two trace terms act as a regularization to choose good inducing sets for $\bm{f}$ and $\bm{g}$, and guard against over-fitting. It is observed that $\mathrm{Tr}[\bm{K}_{nn}^g - \bm{Q}_{nn}^g]$ and $\mathrm{Tr}[\bm{K}_{nn}^f - \bm{Q}_{nn}^f]$ represent the total variance of the training conditionals $p(\bm{g}|\bm{g}_u)$ and $p(\bm{f}|\bm{f}_m)$, respectively. To maximize $F_V$, the traces should be very small, implying that $\bm{f}_m$ and $\bm{g}_u$ are very informative (i.e., sufficient statistics, also called variational compression \cite{hensman2014nested}) for $\bm{f}$ and $\bm{g}$. Particularly, the zero traces indicate that $\bm{f}_m = \bm{f}$ and $\bm{g}_u = \bm{g}$, thus recovering the variational HGP (VHGP)~\cite{titsias2011variational}. Besides, the zero traces imply that the variances of $q(\bm{g}_u)$ equal to that of $p(\bm{g}_u|\bm{y})$.
	\item The KL term is a constraint for rationalising $q(\bm{g}_u)$. It is observed that minimizing the traces only push the variances of $q(\bm{g}_u)$ towards that of $p(\bm{g}_u|\bm{y})$. To let the co-variances of $q(\bm{g}_u)$ rationally approximate that of $p(\bm{g}_u|\bm{y})$, the KL term penalizes $q(\bm{g}_u)$ so that it does not deviate significantly from the prior $p(\bm{g}_u)$.
\end{itemize}

\subsection{Reparameterization and inference}
In order to maximize the ELBO $F_V$ in \eqref{eq_F}, we need to infer $\omega = u + u(u+1)/2$ free variational parameters in $\bm{\mu}_u$ and $\bm{\Sigma}_u$. Assume that $u = 0.01n$, then $\omega$ is larger than $n$ when the training size $n > 2 \times 10^4$, leading to a high-dimensional and non-trivial optimization task.

We observe that the derivatives of $F_V$ w.r.t. $\bm{\mu}_u$ and $\bm{\Sigma}_u$ are
\begin{equation*}
	\begin{aligned}
	\frac{\partial F_V}{\partial \bm{\mu}_{u}} =& \frac{1}{2} (\bm{\Omega}_{nu}^g)^{\mathsf{T}} \bm{\Lambda}_{nn}^{ab} \bm{1} - (\bm{K}_{uu}^g)^{-1} (\bm{\mu}_{u} - \mu_0\bm{1}),  \\
	\frac{\partial F_V}{\partial \bm{\Sigma}_{u}} =& -\frac{1}{4} (\bm{\Omega}_{nu}^g)^{\mathsf{T}} (\bm{\Lambda}_{nn}^{ab} + \bm{I}) \bm{\Omega}_{nu}^g + \frac{1}{2} [\bm{\Sigma}_{u}^{-1} - (\bm{K}_{uu}^g)^{-1}], 
	\end{aligned}
\end{equation*}
where the diagonal matrix $\bm{\Lambda}_{nn}^{ab} = \bm{\Lambda}_{nn}^a + \bm{\Lambda}_{nn}^b$ with $\bm{\Lambda}_{nn}^a = (\bm{\Sigma}_{y}^{-1} \bm{y} \bm{y}^{\mathsf{T}} \bm{\Sigma}_{y}^{-1} - \bm{\Sigma}_{y}^{-1}) \odot \bm{R}_g$, $\bm{\Lambda}_{nn}^b = (\bm{K}_{nn}^f - \bm{Q}_{nn}^f) \odot \bm{R}_g^{-1}$, and the operator $\odot$ being the element-wise product. Hence, it is observed that the optimal $\bm{\mu}_u$ and $\bm{\Sigma}_u^{-1}$ satisfy
\begin{align*}
	\bm{\mu}_u =& 0.5\bm{K}_{un}^g \bm{\Lambda}_{nn}^{ab} \bm{1} + \mu_0\bm{1}, \\
	\bm{\Sigma}_u^{-1} =& 0.5 (\bm{\Omega}_{nu}^g)^{\mathsf{T}} (\bm{\Lambda}_{nn}^{ab} + \bm{I}) \bm{\Omega}_{nu}^g + (\bm{K}_{uu}^g)^{-1}.
\end{align*}
Interestingly, we find that both the optimal $\bm{\mu}_u$ and $\bm{\Sigma}_u^{-1}$ depend on $\bm{\Lambda}_{nn} = 0.5 (\bm{\Lambda}_{nn}^{ab} + \bm{I})$, which is a positive semi-definite diagonal matrix, see the \textit{non-negativity} proof in Appendix~\ref{APP_Non_negativity}. Hence, we re-parameterize $\bm{\mu}_u$ and $\bm{\Sigma}_u^{-1}$ in terms of $\bm{\Lambda}_{nn}$ as
\begin{subequations} \label{eq_mu_sigma_u_para}
	\begin{align}
	\bm{\mu}_u =& \bm{K}_{un}^g (\bm{\Lambda}_{nn}- 0.5\bm{I}) \bm{1} + \mu_0\bm{1}, \label{eq_mu_u_para} \\
	\bm{\Sigma}_u^{-1} =& (\bm{K}_{uu}^g)^{-1} + (\bm{\Omega}_{nu}^g)^{\mathsf{T}} \bm{\Lambda}_{nn} \bm{\Omega}_{nu}^g. \label{eq_Sigma_u_para}
	\end{align}
\end{subequations}
This re-parameterization eases the model inference by (i) reducing the number of variational parameters from $\omega$ to $n$, and (ii) limiting the new variational parameters $\bm{\Lambda}_{nn}$ to be non-negative, thus narrowing the search space. 

So far, the bound $F_V$ depends on the variational parameters $\bm{\Lambda}_{nn}$, the kernel parameters $\bm{\theta}^f$ and $\bm{\theta}^g$, the mean parameter $\mu_0$ for $g$, and the inducing points $\bm{X}_m$ and $\bm{X}_u$. We maximize $F_V$ to infer all these hyperparameters $\bm{\zeta} = \{ \bm{\Lambda}_{nn}, \bm{\theta}^f, \bm{\theta}^g, \bm{X}_m, \bm{X}_u \}$ jointly for model selection. This non-linear optimization task can be solved via conjugate gradient descent (CGD), since the derivatives of $F_V$ w.r.t. these hyperparameters have closed forms, see Appendix~\ref{APP_Derivs_F}.

\subsection{Predictive Distribution}
The predictive distribution $p(y_* | \bm{y}, \bm{x}_*)$ at the test point $\bm{x}_*$ is approximated as
\begin{equation}
q(y_*) = \int p(y_*|f_*, g_*) q(f_*) q(g_*) df_*dg_*.
\end{equation}
As for $q(f_*) = \int p(f_*|\bm{f}_m) q^{\star}(\bm{f}_m) d\bm{f}_m$, we first calculate $q^{\star}(\bm{f}_m) = \mathcal{N}(\bm{f}_m|\bm{\mu}_f^{\star}, \bm{\Sigma}_f^{\star})$ from~\eqref{e1_q*(fm)} as
\begin{align} \label{eq_q*(fm)_closedform}
\bm{\mu}_f^{\star} = \bm{K}_{mm}^f\bm{K}_{R}^{-1}\bm{K}_{mn}^f\bm{R}_g^{-1}\bm{y}, \quad \bm{\Sigma}_f^{\star} = \bm{K}_{mm}^f\bm{K}_{R}^{-1}\bm{K}_{mm}^f,
\end{align}
where $\bm{K}_{R} = \bm{K}_{mn}^f\bm{R}_g^{-1}\bm{K}_{nm}^f + \bm{K}_{mm}^f$. Using $q^{\star}(\bm{f}_m)$, we have $q(f_*) = \mathcal{N}(f_*|\mu_{f*}, \sigma_{f*}^2)$
with
\begin{subequations} \label{eq_mu_f*,s2_f*}
	\begin{align}
	\mu_{f*} =& \bm{k}_{*m}^f\bm{K}_{R}^{-1}\bm{K}_{mn}^f\bm{R}_g^{-1}\bm{y}, \label{eq_mu_f} \\
	\sigma_{f*}^2 = & k_{**}^f - \bm{k}_{*m}^f(\bm{K}_{mm}^f)^{-1}  \bm{k}_{m*}^f + \bm{k}_{*m}^f \bm{K}_{R}^{-1} \bm{k}_{m*}^f. \label{eq_s2_f}
	\end{align}
\end{subequations}
It is interesting to find in~\eqref{eq_s2_f} that the correction term $\bm{k}_{*m}^f \bm{K}_{R}^{-1} \bm{k}_{m*}^f$ contains the heteroscedasticity information from the noise term $\bm{R}_g$. Hence, $q(f_*)$ produces \textit{heteroscedastic} variances over the input domain, see an illustration example in Fig.~\ref{Fig_DVSHGP_toy}(b). The heteroscedastic $\sigma^2_{f}(\bm{x}_*)$ (i) eases the learning of $g$, and (ii) plays as an auxiliary role, since the heteroscedasticity is mainly explained by $g$. Also, our VSHGP is believed to produce a better prediction mean $\mu_{f}(\bm{x}_*)$ through the interaction between $f$ and $g$ in~\eqref{eq_mu_f}.\footnote{This happens when $g$ is learned well, see the numerical experiments below.}

Similarly, we have the predictive distribution $q(g_*) = \int p(g_*|\bm{g}_{u}) q(\bm{g}_{u}) d\bm{g}_{u} = \mathcal{N}(g_*|\mu_{g*}, \sigma_{g*}^2)$
where, given $\bm{K}_{\Lambda} = \bm{K}_{un}^g\bm{\Lambda}_{nn}^{-1}\bm{K}_{nu}^g + \bm{K}_{uu}^g$,
\begin{subequations} \label{eq_mu_g*,s2_g*}
	\begin{align}
	\mu_{g*} =& \bm{k}_{*u}^g (\bm{K}_{uu}^g)^{-1} (\bm{\mu}_{u} -\mu_0\bm{1}) + \mu_0 \bm{1}, \\
	\sigma^2_{g*} =& k_{**}^g - \bm{k}_{*u}^g (\bm{K}_{uu}^g)^{-1} \bm{k}_{u*}^g + \bm{k}_{*u}^g \bm{K}_{\Lambda}^{-1} \bm{k}_{u*}^g.
	\end{align}
\end{subequations}

Finally, using the posteriors $q(f_*)$ and $q(g_*)$, and the likelihood $p(y_*|f_*,g_*) = \mathcal{N}(y_*| f_*, e^{g_*})$, we have
\begin{equation} \label{eq_q(y_*)}
\begin{aligned}
q(y_*) = \int \mathcal{N}(y_*|\mu_{f*}, e^{g_*} + \sigma_{f*}^2) \mathcal{N}(g_*|\mu_{g*}, \sigma_{g*}^2) dg_*,
\end{aligned}
\end{equation}
which is intractable and \textit{non-Gaussian}. However, the integral can be approximated up to several digits using the Gauss-Hermite quadrature, resulting in the mean and variance as \cite{titsias2011variational}
\begin{align} \label{eq_mean_s2}
	\mu_* = \mu_{f*}, \quad \sigma^2_* = \sigma_{f*}^2 + e^{\mu_{g*} + \sigma_{g*}^2/2}. 
\end{align}
In the final prediction variance, the variance $\sigma^2_{f*}$ represents the uncertainty about $f$ due to data density, and it approaches zero with increasing $n$; the exponential term implies the intrinsic heteroscedastic noise uncertainty.

It is notable that the unifying VSHGP includes VHGP \cite{titsias2011variational} and variational sparse GP (VSGP) \cite{titsias2009variational} as special cases: when $\bm{f}_m = \bm{f}$ and $\bm{g}_u = \bm{g}$, VSHGP recovers VHGP; when $q(\bm{g}_u) = p(\bm{g}_u)$, i.e., we are now facing a homoscedastic regression task, VSHGP regenerates to VSGP.

Overall, by introducing inducing sets for both $\bm{f}$ and $\bm{g}$, VSHGP is equipped with the means to handle large-scale heteroscedastic regression. However, (i) the current time complexity $\mathcal{O}(nm^2+nu^2)$, which is linear with training size, makes VSHGP still unaffordable for, e.g., millions of data points; and (ii) as a global approximation, the capability of VSHGP is limited by the \textit{small} and \textit{global} inducing sets.

To this end, we introduce below two strategies to further improve the scalability and capability of VSHGP.

\section{Stochastic VSHGP} 
\label{sec_SVSHGP}
To further improve the \textit{scalability} of VSHGP, the variational distribution $q(\bm{f}_m) = \mathcal{N}(\bm{f}_m|\bm{\mu}_m, \bm{\Sigma}_m)$ is re-introduced to use the original bound $F = \int q(\bm{z}) \log \frac{p(\bm{z}, \bm{y})}{q(\bm{z})} d\bm{z}$ in~\eqref{eq_VB}. Given the factorized likelihood $p(\bm{y}|\bm{f},\bm{g}) = \prod_{i=1}^N p(y_i|f_i,g_i)$, the ELBO $F$ is
\begin{equation} \label{eq_svi_F}
\begin{aligned}
F =& \sum_{i=1}^n \mathbb{E}_{q(f_i)q(g_i)}[\log p(y_i|f_i, g_i)]- \mathrm{KL}[q(\bm{f}_m) || p(\bm{f}_m)] \\
&- \mathrm{KL}[q(\bm{g}_u) || p(\bm{g}_u)],
\end{aligned}
\end{equation}
where the first expectation term is expressed as
\begin{align*}
=:&\sum_{i=1}^n \left[ \log \mathcal{N}(y_i|[\bm{\mu}_f]_i, [\bm{R}_g]_{ii}) - \frac{1}{4}[\bm{\Sigma}_g]_{ii} - \frac{1}{2}[\bm{\Sigma}_f\bm{R}_g^{-1}]_{ii} \right],
\end{align*}
with $\bm{\mu}_{f} = \bm{\Omega}_{nm}^f \bm{\mu}_{m}$ and $\bm{\Sigma}_{f} = \bm{K}_{nn}^f - \bm{Q}_{nn}^f + \bm{\Omega}_{nm}^f \bm{\Sigma}_{m} (\bm{\Omega}_{nm}^f)^{\mathsf{T}}$. 

The new $F$ is a relaxed version of $F_V$ in~\eqref{eq_F}. It is found that the derivatives satisfy
\begin{equation} \label{eq_partialF_mu_m_s_m}
\begin{aligned}
\frac{\partial F}{\partial \bm{\mu}_{m}} =& (\bm{\Omega}_{nm}^f)^{\mathsf{T}} \bm{R}_{g}^{-1} (\bm{y} - \bm{\Omega}_{nm}^f \bm{\mu}_m) - (\bm{K}_{mm}^f)^{-1} \bm{\mu}_m, \\
\frac{\partial F}{\partial \bm{\Sigma}_{m}} =& -\frac{1}{2} (\bm{\Omega}_{nm}^f)^{\mathsf{T}} \bm{R}_{g}^{-1} \bm{\Omega}_{nm}^f + \frac{1}{2} (\bm{\Sigma}_{m}^{-1} - (\bm{K}_{mm}^f)^{-1}).   
\end{aligned}
\end{equation}
Let the gradients be zeros, we recover the optimal solution $q^{\star}(\bm{f}_m)$ in~\eqref{eq_q*(fm)_closedform}, indicating that $F_V \ge F$ with the equality at $q(\bm{f}_m) = q^{\star}(\bm{f}_m)$. 

The scalability is improved by $F$ through the first term in the right-hand side of~\eqref{eq_svi_F}, which factorizes over data points. The sum form allows using efficient stochastic gradient descent (SGD), e.g., Adam~\cite{kingma2014adam}, with mini-batch mode for big data. Specifically, we choose a random subset $\mathcal{B} \subseteq \{1, \cdots, n\}$ to have an unbiased estimation of $F$ as
\begin{equation}
\begin{aligned}
\widetilde{F} = & \frac{n}{|\mathcal{B}|}\sum_{i \in \mathcal{B}} \mathbb{E}_{q(f_i)q(g_i)}[\log p(y_i|f_i, g_i)] \\
&- \mathrm{KL}[q(\bm{f}_m) || p(\bm{f}_m)] - \mathrm{KL}[q(\bm{g}_u) || p(\bm{g}_u)],
\end{aligned}
\end{equation}
where $|\mathcal{B}| \ll n$ is the mini-batch size. More efficiently, since the two variational distributions are defined in terms of KL divergence, we could optimize them along the \textit{natural} gradients instead of the Euclidean gradients, see Appendix~\ref{sec_natural_gradients}.

Finally, the predictions of the stochastic VSHGP (SVSHGP) follow~\eqref{eq_mean_s2}, with the predictions replaced as
\begin{equation*}
	\begin{aligned}
	\mu_{f*} =& \bm{k}_{*m}^f (\bm{K}_{mm}^f)^{-1} \bm{\mu}_{m}, \nonumber \\
	\sigma^2_{f*} =& k_{**}^f - \bm{k}_{*m}^f (\bm{K}_{mm}^f)^{-1} (\bm{K}_{mm}^f - \bm{\Sigma}_{m}) (\bm{K}_{mm}^f)^{-1} \bm{k}_{m*}^f, \nonumber
	\end{aligned}
\end{equation*}
and
\begin{equation*}
	\begin{aligned}
	\mu_{g*} =& \bm{k}_{*u}^g (\bm{K}_{uu}^g)^{-1} (\bm{\mu}_{u} -\mu_0\bm{1}) + \mu_0 \bm{1}, \\
	\sigma^2_{g*} =& k_{**}^g - \bm{k}_{*u}^g (\bm{K}_{uu}^g)^{-1} (\bm{K}_{uu}^g - \bm{\Sigma}_{u}) (\bm{K}_{uu}^g)^{-1} \bm{k}_{u*}^g.
	\end{aligned}
\end{equation*}

Compared to the deterministic VSHGP, the stochastic variant greatly reduces the time complexity from $\mathcal{O}(nm^2+nu^2)$ to $\mathcal{O}(|\mathcal{B}|m^2+|\mathcal{B}|u^2+m^3+u^3)$, at the cost of requiring many more optimization efforts in the enlarged probabilistic space.\footnote{Compared to VSHGP, the SVSHGP cannot re-parameterize $\bm{\mu}_u$ and $\bm{\Sigma}_u$, and has to infer $m+m(m+1)$ more variational parameters in $\bm{\mu}_m$ and $\bm{\Sigma}_m$.} Besides, the capability of SVSHGP akin to VSHGP is still limited to the finite number of global inducing points.

\section{Distributed VSHGP}
\label{sec_DVSHGP}
To further improve the \textit{scalability} and \textit{capability} of VSHGP via \textit{many} inducing points, the distributed variant named DVSHGP proposes to combine VSHGP with local approximations, e.g., the Bayesian committee machine (BCM) \cite{tresp2000bayesian, deisenroth2015distributed}, to enable distributed learning and capture local variety.

\subsection{Training experts with hybrid parameters}
We first partition the training data $\mathcal{D}$ into $M$ subsets $\mathcal{D}_i = \{ \bm{X}_i, \bm{y}_i \}$, $1 \le i \le M$. Then, we train a VSHGP expert $\mathcal{M}_i$ on $\mathcal{D}_i$ by using the relevant inducing sets $\bm{X}_{m_i}$ and $\bm{X}_{u_i}$. Particularly, to obtain computational gains, an independence assumption is posed for all the experts $\{ \mathcal{M}_i \}_{i=1}^M$ such that $\log p(\bm{y}; \bm{X}, \bm{\zeta})$ is decomposed into the sum of $M$ individuals
\begin{equation} \label{eq_p(y)_decompose}
\log p(\bm{y} ; \bm{X}, \bm{\zeta}) \approx \sum_{i=1}^M \log p(\bm{y}_i ; \bm{X}_i, \bm{\zeta}_i) \ge \sum_{i=1}^M F_{V_i},
\end{equation}
where $\bm{\zeta}_i$ is the hyperparameters in $\mathcal{M}_i$, and $F_{V_i} = F(q(\bm{g}_{u_i}))$ is the ELBO of $\mathcal{M}_i$. The factorization~\eqref{eq_p(y)_decompose} calculates the inversions efficiently as $(\bm{K}_{mm}^{f})^{-1} \approx \mathrm{diag}[\{(\bm{K}^{f}_{m_im_i})^{-1}\}_{i=1}^M]$ and $(\bm{K}_{mm}^{g})^{-1} \approx \mathrm{diag}[\{(\bm{K}^{g}_{m_im_i})^{-1}\}_{i=1}^M]$.

We train these VSHGP experts with \textit{hybrid} parameters. Specifically, the BCM-type aggregation requires sharing the priors $p(f_*)$ and $p(g_*)$ over experts. That means, we should share the hyperparameters including $\bm{\theta}^f$, $\bm{\theta}^g$ and $\mu_0$ across experts. These \textit{global} parameters are beneficial for guarding against over-fitting~\cite{deisenroth2015distributed}, at the cost of however degrading the capability. Hence, we leave the variational parameters $\bm{\Lambda}_{n_i n_i}$ and the inducing points $\bm{X}_{m_i}$ and $\bm{X}_{u_i}$ for each expert to infer them individually. These \textit{local} parameters improve capturing local variety by (i) pushing $q(\bm{g}_{u_i})$ towards the posterior $p(\bm{g}_{u_i}|\bm{y}_i)$ of $\mathcal{M}_i$, and (ii) using many inducing points.

Besides, because of the local parameters, we prefer partitioning the data into \textit{disjoint} experts rather than \textit{random} experts like \cite{deisenroth2015distributed}. The disjoint partition using clustering techniques produces local and separate experts which are desirable for learning the relevant local parameters. In contrast, the random partition, which assigns points randomly to the subsets, provides global and overlapped experts which are difficult to well estimate the local parameters. For instance, when DVSHGP uses random experts on the toy example below, it fails to capture the heteroscedastic noise.

Finally, suppose that each expert has the same training size $n_0=n/M$, the training complexity for an expert is $\mathcal{O}(n_0m_0^2+n_0u_0^2)$, where $m_0$ is the inducing size for $\bm{f}_i$ and $u_0$ the inducing size for $\bm{g}_i$. Due to the $M$ local experts, DVSHGP naturally offers parallel/distributed training, hence reducing the time complexity of VSHGP with a factor ideally close to the number of machines when $m_0=m$ and $u_0=u$.

\subsection{Aggregating experts' predictions}
For each VSHGP expert $\mathcal{M}_i$, we obtain the predictive distribution $q_i(y_*)$ with the means $\{\mu_{f_i*}, \mu_{g_i*}, \mu_{i*} \}$ and variances $\{ \sigma^2_{f_i*}, \sigma^2_{g_i*}, \sigma^2_{i*} \}$. Thereafter, we combine the experts' predictions together to perform the final prediction by, for example the robust BCM (RBCM) aggregation, which naturally supports distributed/parallel computing~\cite{deisenroth2015distributed, ingram2010parallel}.

The key to the success of aggregation is that we do not directly combine the experts' predictions $\{ \mu_{i*}, \sigma^2_{i*} \}_{i=1}^M$. This is because (i) the RBCM aggregation of $\{q_i(y_*)\}_{i=1}^M$ produces an inconsistent prediction variance with increasing $n$ and $M$~\cite{liu2018generalized}; and (ii) the predictive distribution $q_i(y_*)$ in~\eqref{eq_q(y_*)} is non-Gaussian.\footnote{Note that the generalized RBCM strategy~\cite{liu2018generalized}, which provides consistent predictions, however is not favored by our DVSHGP with local parameters, since it requires (i) the experts' predictions to be Gaussian and (ii) an additional global base expert.} To have a meaningful prediction variance, which is crucial for heteroscedastic regression, we perform the aggregation for the latent functions $f$ and $g$, respectively. This is because the prediction variances of $f$ and $g$ approach zeros with increasing $n$. This agrees with the property of RBCM.

We first have the aggregated predictive distribution for $f_*$, given the prior $p(f_*)=\mathcal{N}(f_*| 0, \sigma_{f**}^{2} \triangleq k^f_{**})$, as
\begin{equation}
p_{\mathcal{A}}(f_*|\bm{y}, \bm{x}_*) = \frac{\prod_{i=1}^M [q_{i}(f_*)]^{w^{f}_{i*}}}{[p(f_*)]^{\sum_i w^{f}_{i*}-1}},
\end{equation}
with the mean and variance expressed respectively, as
\begin{align*}
	\mu_{f_\mathcal{A}*} &= \sigma_{f_\mathcal{A}*}^2 \sum_{i=1}^M w^{f}_{i*} \sigma_{f_i*}^{-2} \mu_{f_i*},  \\
	\sigma_{f_\mathcal{A}*}^{-2} &= \sum_{i=1}^M w^f_{i*} \sigma_{f_i*}^{-2} + \left(1-\sum_{i=1}^M w^f_{i*} \right)\sigma_{f**}^{-2}, 
\end{align*}
where the weight $w^f_{i*} \ge 0$ represents the contribution of $\mathcal{M}_i$ at $\bm{x}_*$ for $f$, and is defined as the difference in the differential entropy between the prior $p(f_*)$ and the posterior $q_i(f_*)$ as $w^f_{i*} = 0.5(\log\sigma_{f**}^{2} - \log\sigma_{f_i*}^2)$. Similarly, for $g$ which explicitly considers a prior mean $\mu_0$, the aggregated predictive distribution is
\begin{equation}
p_{\mathcal{A}}(g_*|\bm{y}, \bm{x}_*) = \frac{\prod_{i=1}^M [q_{i}(g_*)]^{w^g_{i*}}}{[p(g_*)]^{\sum_i w^g_{i*}-1}},
\end{equation}
with the mean and variance, expressed respectively as
\begin{align*}
	\mu_{g_\mathcal{A}*} =& \sigma_{g_\mathcal{A}*}^2 \left[ \sum_{i=1}^M w^g_{i*} \sigma_{g_i*}^{-2} \mu_{g_i*} + \left(1- \sum_{i=1}^M w^g_{i*} \right) \sigma_{g**}^{-2} \mu_0 \right], \label{Eq_mug_RBCM} \\
	\sigma_{g_\mathcal{A}*}^{-2} =& \sum_{i=1}^M w^g_{i*} \sigma_{g_i*}^{-2} + \left(1-\sum_{i=1}^M w^g_{i*} \right)\sigma_{g**}^{-2},
\end{align*}
where $\sigma_{g**}^{-2}$ is the prior precision of $g_*$, and $w^g_{i*} = 0.5(\log\sigma_{g**}^{2} - \log\sigma_{g_i*}^2)$ is the weight of $\mathcal{M}_i$ at $\bm{x}_*$ for $g$.

\begin{figure}[!t] 
	\centering
	\includegraphics[width=3.0in]{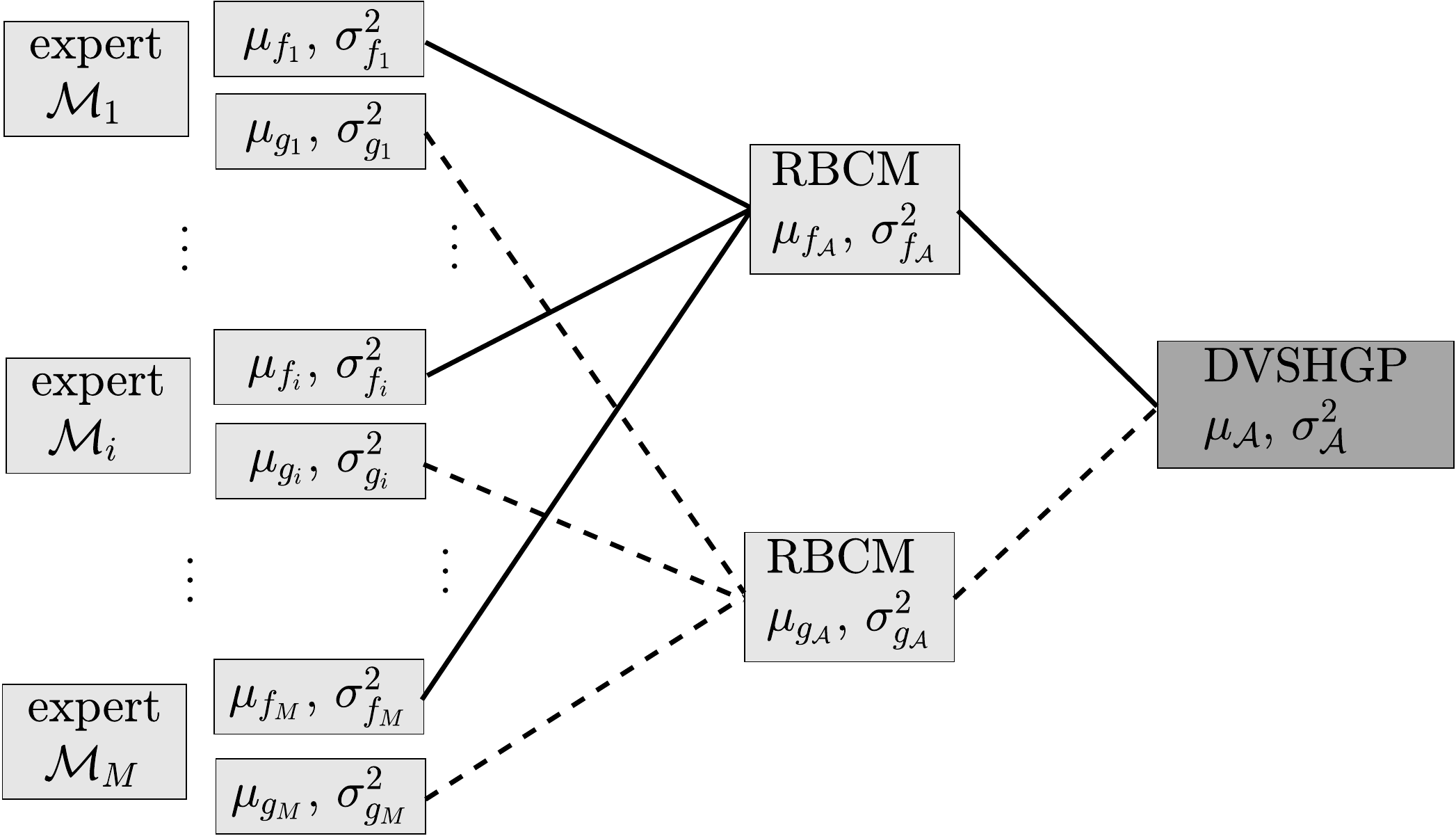}
	\caption{Hierarchy of the proposed DVSHGP model.}
	\label{Fig_DVSHGP_flow} 
\end{figure}

Thereafter, as shown in Fig.~\ref{Fig_DVSHGP_flow}, the final predictions akin to~\eqref{eq_mean_s2} are combined as
\begin{align}
	\mu_{\mathcal{A}*} = \mu_{f_\mathcal{A}*}, \quad \sigma^2_{\mathcal{A}*} = \sigma_{f_\mathcal{A}*}^2 + e^{\mu_{g_\mathcal{A}*} + \sigma_{g_\mathcal{A}*}^2/2}. 
\end{align}
The hierarchical and localized computation structure enables (i) large-scale regression via distributed computations, and (ii) flexible approximation of slow-/quick-varying features by local experts and many inducing points (up to the training size $n$).

\begin{figure}[!htb] 
	\centering
	\includegraphics[width=3.2in]{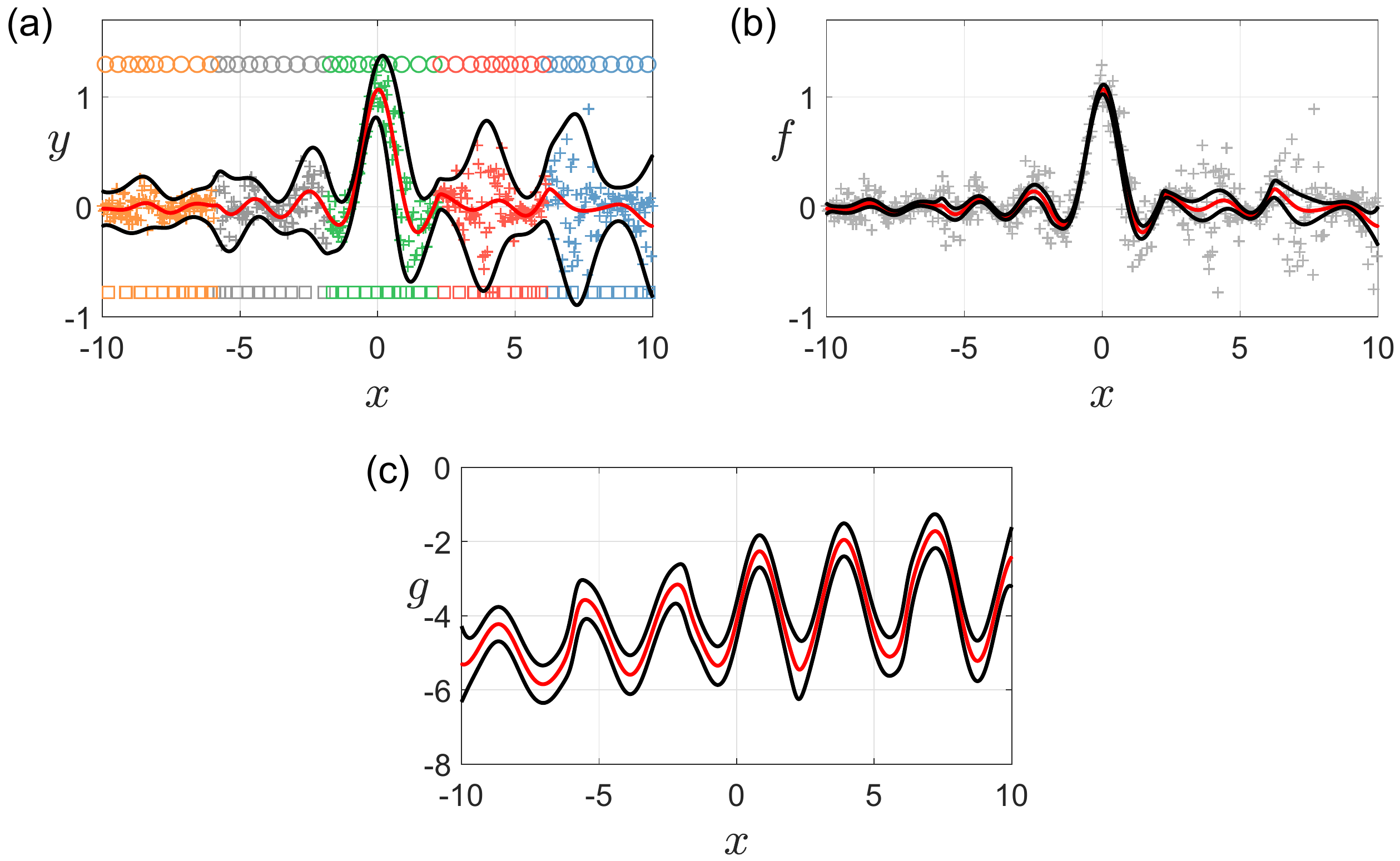}
	\caption{Illustration of DVSHGP on the toy example. The crosses marked in different colors in (a) represent $M=5$ subsets. The top circles and bottom squares marked in different colors represent the optimized positions of inducing points for $\{\bm{f}_i\}_{i=1}^M$ and $\{\bm{g}_i\}_{i=1}^M$, respectively. The red curves present the prediction mean, whereas the black curves represent 95\% confidence interval of the prediction mean.}
	\label{Fig_DVSHGP_toy} 
\end{figure}

Finally, we illustrate the DVSHGP on a heteroscedastic toy example expressed as
\begin{equation} \label{eq_toy}
y(x) = \mathrm{sinc}(x) + \epsilon, \quad x \in [-10,10],
\end{equation}
where $\epsilon = \mathcal{N}(0, \sigma_{\epsilon}^2(x))$ and $\sigma_{\epsilon}(x) = 0.05 + 0.2(1+\sin(2x)) / (1+e^{-0.2x})$. We draw 500 training points from~\eqref{eq_toy}, and use the $k$-means technique to partition them into five disjoint subsets. We then employ ten inducing points for both $\bm{f}_i$ and $\bm{g}_i$ of the VSHGP expert $\mathcal{M}_i$, $1 \le i \le 5$. Fig.~\ref{Fig_DVSHGP_toy} shows that (i) DVSHGP can efficiently employ up to 100 inducing points for modeling through five local experts, and (ii) DVSHGP successfully describes the underlying function $f$ and the heteroscedastic log noise variance $g$.

\section{Discussions regarding implementation}
\subsection{Implementation of DVSHGP}
As for DVSHGP, we should infer (i) the global parameters including the kernel parameters $\bm{\theta}^f$ and $\bm{\theta}^g$, and the mean $\mu_0$; and (ii) the local parameters including the variational parameters $\bm{\Lambda}_{n_i n_i}$ and the inducing parameters $\bm{X}_{m_i}$ and $\bm{X}_{u_i}$, for local experts $\{\mathcal{M}_i\}_{i=1}^M$.

Notably, the variational parameters $ \bm{\Lambda}_{n_i n_i}$ are crucial for the success of DVSHGP, since they represent the heteroscedasticity. To learn the variational parameters well, there are two issues: (i) how to initialize them and (ii) how to optimize them. As for initialization, let us focus on VSHGP, which is the foundation for the experts in DVSHGP. It is observed in~\eqref{eq_mu_sigma_u_para} that $\bm{\Lambda}_{nn}$ directly determines the initialization of $q(\bm{g}_u) = \mathcal{N}(\bm{g}_u|\bm{\mu}_u, \bm{\Sigma}_u)$. Given the prior $\bm{g}_u \sim \mathcal{N}(\bm{g}_{u}|\mu_0 \bm{1}, \bm{K}_{uu}^g)$, we intuitively place a prior mean $\mu_0 \bm{1}$ on $\bm{\mu}_u$, resulting in $\bm{\Lambda}_{nn} = 0.5 \bm{I}$. In contrast, if we initialize $[\bm{\Lambda}_{nn}]_{ii}$ with a value larger or smaller than $0.5$, the cumulative term $\bm{K}_{un}^g (\bm{\Lambda}_{nn}- 0.5\bm{I}) \bm{1}$ in~\eqref{eq_mu_u_para} becomes far away from zero with increasing $n$, leading to improper prior mean for $\bm{\mu}_u$. As for optimization, compared to standard GP, our DVSHGP needs to additionally infer $n$ variational parameters and $M(m_0+u_0)d$ inducing parameters, which greatly enlarge the parameter space and increase the optimization difficulty. Hence, we use an alternating strategy where we first optimize the variational parameters individually to capture the heteroscedasticity roughly, followed by learning all the hyperparameters jointly. 

\begin{figure}[!t] 
	\centering
	\includegraphics[width=3.2in]{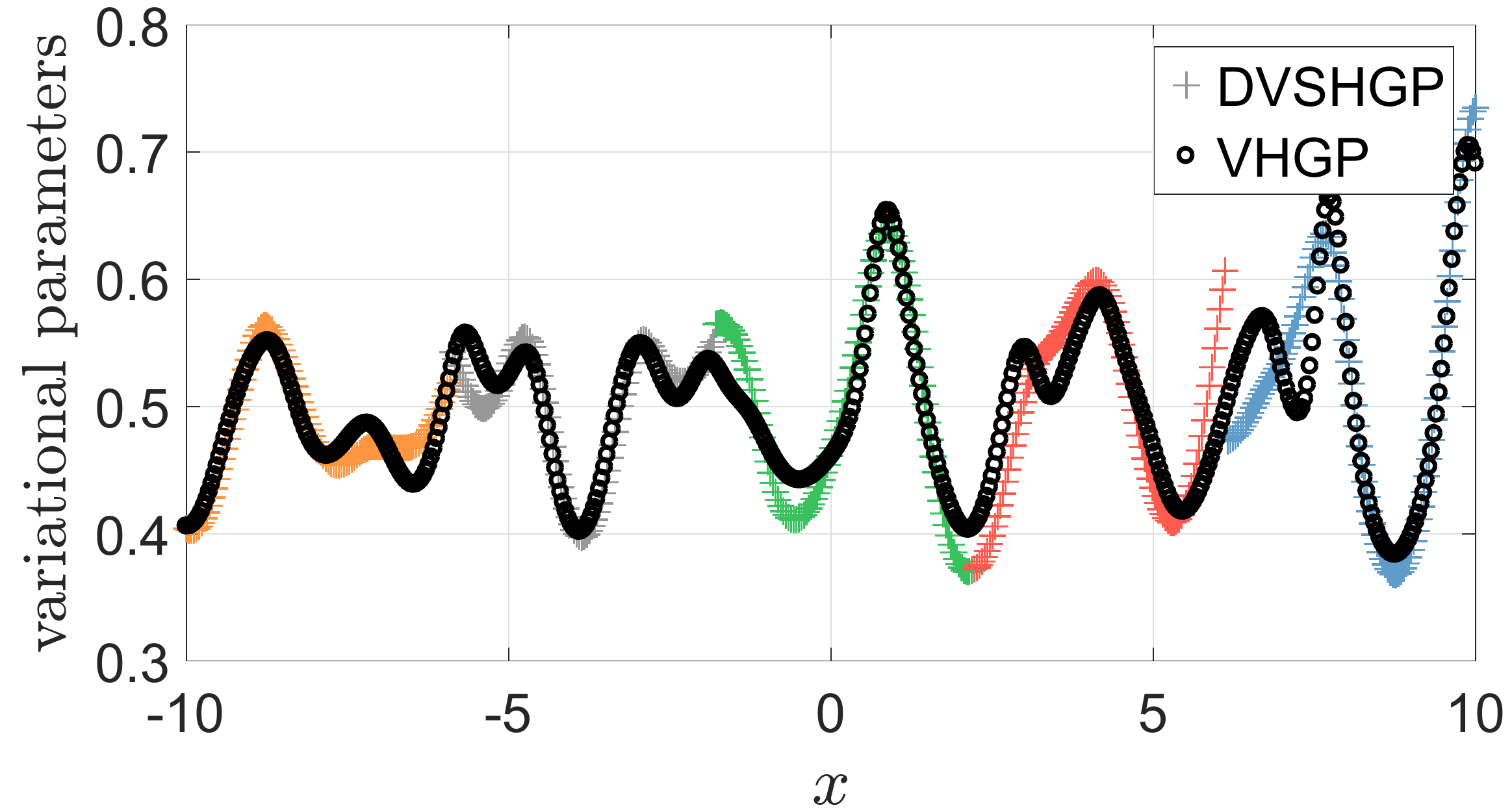}
	\caption{The variational parameters learnt respectively by DVSHGP and VHGP on the toy problem. The crosses marked in different colors represent the variational parameters inferred for the five local experts of DVSHGP.}
	\label{Fig_toy_VP_DVSHGP_VHGP} 
\end{figure}

Fig.~\ref{Fig_toy_VP_DVSHGP_VHGP} depicts the inferred variational parameters varying over training points by DVSHGP and the original VHGP \cite{titsias2011variational}, respectively, on the toy problem. It turns out that the variational parameters estimated by DVSHGP (i) generally agree with that of VHGP, and (ii) showcase local characteristics that are beneficial for describing local variety.

\subsection{Implementation of SVSHGP}
As for SVSHGP, to effectively infer the variational parameters in $q(\bm{f}_m)$ and $q(\bm{g}_u)$, we adopt the natural gradient descent (NGD), which however should carefully tune the step parameter $\gamma$ defined in Appendix~\ref{sec_natural_gradients}. For the Gaussian likelihood, the optimal solution is $\gamma=1.0$, since taking an unit step is equivalent to performing a VB update~\cite{hensman2013gaussian}. But for the stochastic case, empirical results suggest that $\gamma$ should be gradually increased to some fixed value. Hence, we follow the schedule in~\cite{salimbeni2018natural}: take $\gamma_{\mathrm{initial}} = 0.0001$ and log-linearly increase $\gamma$ to $\gamma_{\mathrm{final}} = 0.1$ over five iterations, and then keep $\gamma_{\mathrm{final}}$ for the remaining iterations.
	
Thereafter, we employ a hybrid strategy, called NGD+Adam, for optimization. Specifically, we perform a step of NGD on the variational parameters with the aforementioned $\gamma$ schedule, followed by a step of Adam on the remaining hyperparameters with a fixed step $\gamma = 0.01$.

\begin{figure}[!t] 
	\centering
	\includegraphics[width=2.5in]{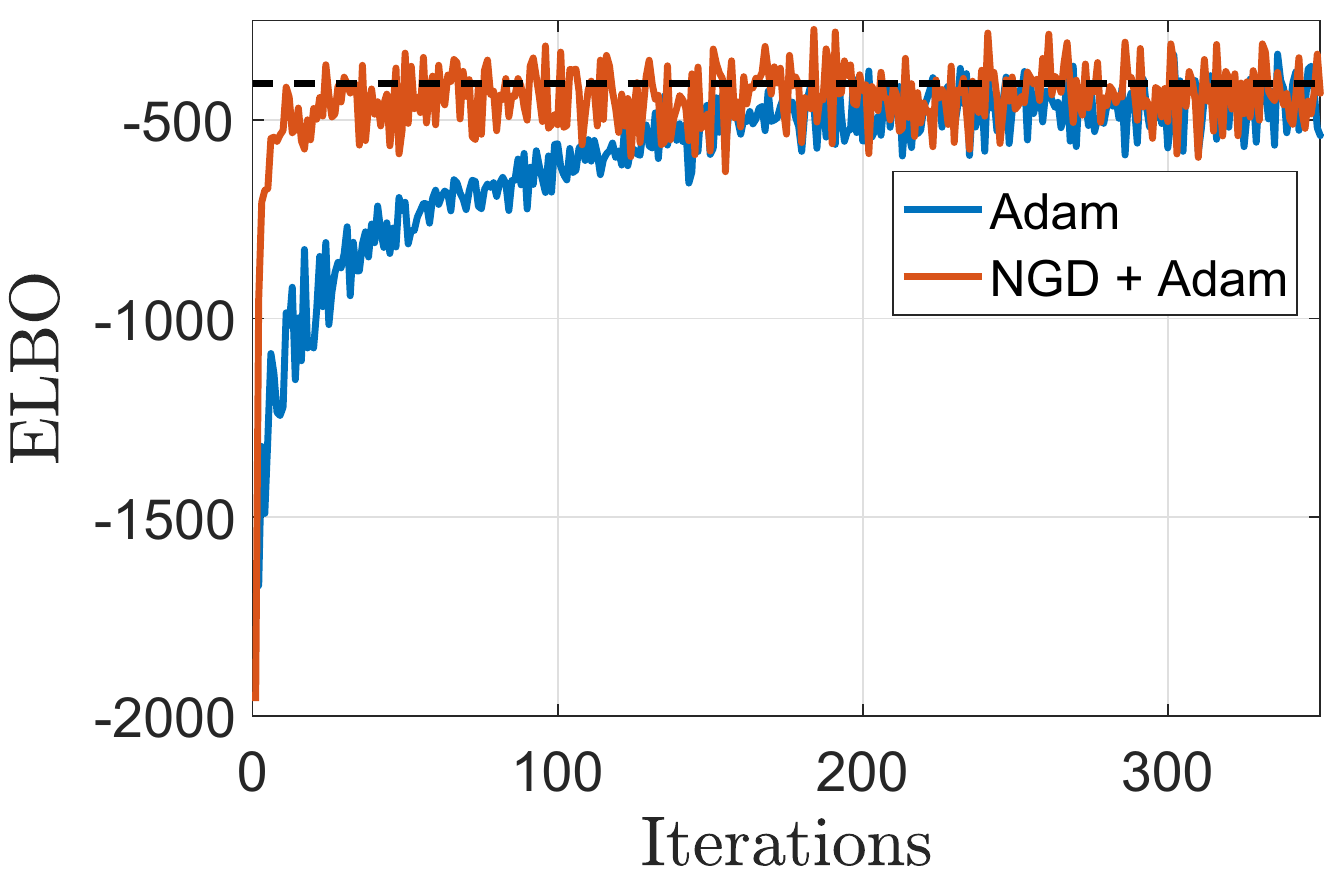}
	\caption{Illustration of SVSHGP using Adam and NGD+Adam respectively  on the toy exmaple. The dash line represents the final ELBO of VSHGP.}
	\label{Fig_toy_SVSHGP_adam_ngd} 
\end{figure}

Fig.~\ref{Fig_toy_SVSHGP_adam_ngd} depicts the convergence histories of SVSHGP using Adam and NGD+Adam respectively on the toy example. We use $m=u=20$ inducing points and a mini-batch size of $|\mathcal{B}|=50$. As the ground truth, the final ELBO obtained by VSHGP is provided. It is observed that (i) the NGD+Adam converges faster than the pure Adam, and (ii) the stochastic optimizers finally approach the solution of CGD used in VSHGP.

\subsection{DVSHGP vs. (S)VSHGP}
Compared to the global (S)VSHGP, the performance of DVSHGP is enhanced by \textit{many} inducing points and the localized experts with \textit{individual} variational and inducing parameters, resulting in the capability of capturing quick-varying features. To verify this, we apply DVSHGP and (S)VSHGP to the time-series \textit{solar irradiance} dataset \cite{hensman2018variational} which contains quick-varying and heteroscedastic features.

\begin{figure}[!htb] 
	\centering
	\includegraphics[width=3.5in]{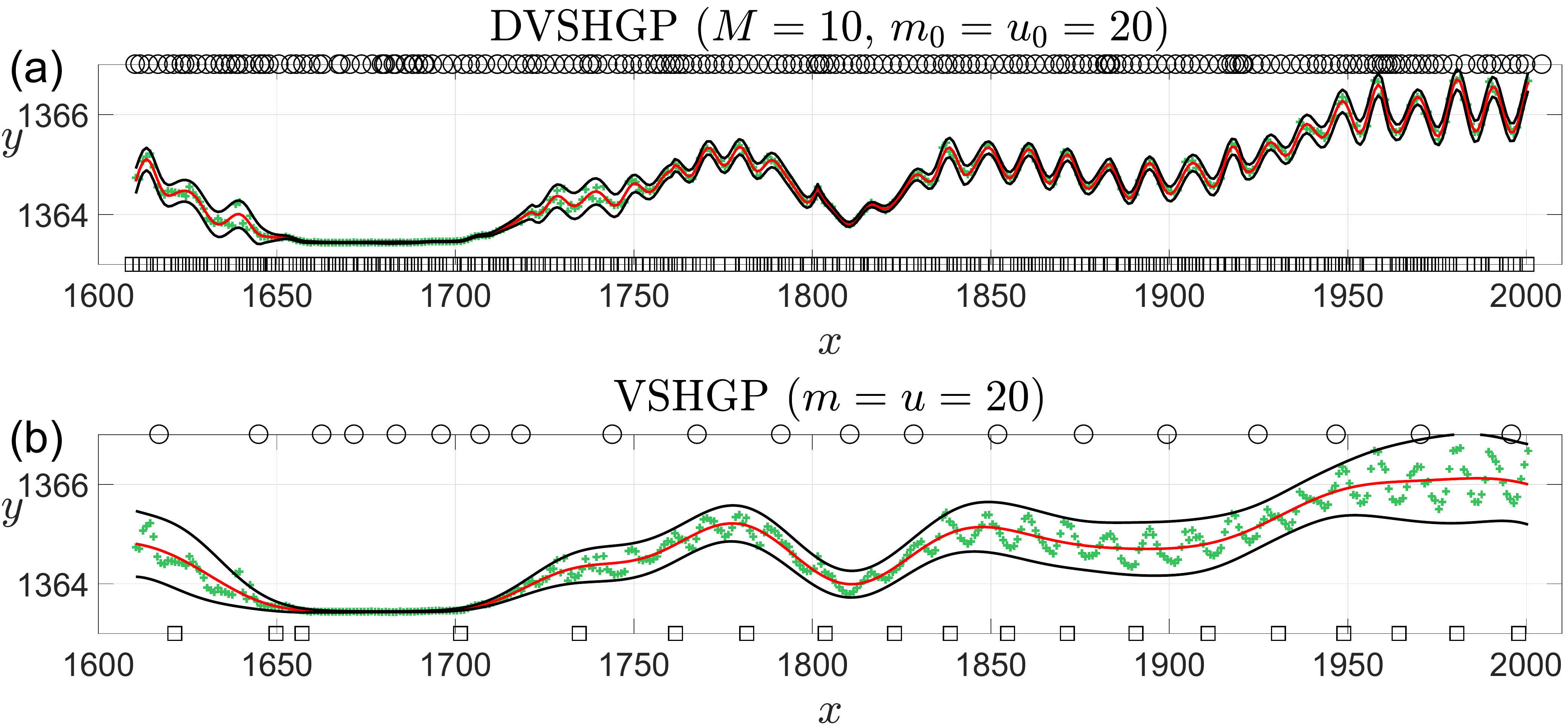}
	\caption{Comparison of DVSHGP and VSHGP on the \textit{solar  irradiance} dataset.}
	\label{Fig_solar_DVSHGP_VHGP} 
\end{figure}

In the comparison, DVSHGP employs the $k$-means technique to partition the 391 training points into $M=10$ subsets, and uses $m_0 = u_0 = 20$ inducing points for each expert; (S)VSHGP employs $m=u=20$ inducing points. Particularly, we initialize the length-scales in the SE kernel~\eqref{eq_SE_Kernel} with a pretty small value of 5.0 for $k^f$ and $k^g$ for (S)VSHGP on this quick-varying dataset. Fig.~\ref{Fig_solar_DVSHGP_VHGP} shows that (i) DVSHGP captures the quick-varying and heteroscedastic features successfully via local experts and many inducing points; (ii) (S)VSHGP however fails due to the small set of global inducing points.\footnote{Since VSHGP and SVSHGP show similar predictions on this dataset, we only illustrate the VSHGP predictions here.}

\section{Numerical experiments} \label{sec_results}
This section verifies the proposed DVSHGP and SVSHGP against existing scalable HGPs on a synthetic dataset and four real-world datasets. The comparison includes (i) GPVC~\cite{almosallam2016gpz}, (ii) the distributed variant of PIC (dPIC)~\cite{hoang2016distributed}, (iii) FITC~\cite{snelson2006sparse}, and (iv) the SoD based empirical HGP (EHSoD)~\cite{urban2015sensor}. Besides, the comparison also employs the homoscedastic VSGP~\cite{titsias2009variational} and RBCM~\cite{deisenroth2015distributed} to showcase the benefits brought by the consideration of heteroscedasticity.

We implement DVSHGP, FITC, EHSoD, VSGP and RBCM by the GPML toolbox~\cite{rasmussen2010gaussian}, and implement SVSHGP by the GPflow toolbox~\cite{matthews2017gpflow}; we use the published GPVC codes available at \url{https://github.com/OxfordML/GPz} and the dPIC codes available at \url{https://github.com/qminh93/dSGP_ICML16}. They are executed on a personal computer with four 3.70 GHz cores and 16 GB RAM for the synthetic and three medium-sized datasets, and on a Linux workstation with eight 3.20 GHz cores and 32GB memory for the large-scale dataset. 

All the GPs employ the SE kernel in~\eqref{eq_SE_Kernel}. Normalization is performed for both $\bm{X}$ and $\bm{y}$ to have zero mean and unit variance before training. Finally, we use $n_*$ test points $\{ \bm{X}_*, \bm{y}_* \}$ to assess the model accuracy by the standardized mean square error (SMSE) and the mean standardized log loss (MSLL)~\cite{rasmussen2006gaussian}. The SMSE quantifies the discrepancy between the predictions and the exact observations. Particularly, it equals to one when the model always predicts the mean of $\bm{y}$. Moreover, the MSLL quantifies the predictive distribution, and is negative for better models. Particularly, it equals to zero when the model always predicts the mean and variance of $\bm{y}$.

\subsection{Synthetic dataset}
We  employ a 2D version of the toy example~\eqref{eq_toy} as
\begin{equation*}
y(\bm{x}) = f(\bm{x}) + \epsilon(\bm{x}), \quad \bm{x} \in [-10, 10]^2,
\end{equation*}
with highly nonlinear latent function $f(\bm{x}) = \mathrm{sinc}(0.1 x_1 x_2)$ and noise $\epsilon(\bm{x}) = \mathcal{N}(0, \sigma_{\epsilon}^2(0.1 x_1 x_2))$. We randomly generate 10,000 training points and evaluate the model accuracy on 4,900 grid test points. We generate ten instances of the training data such that each model is repeated ten times.

We have $M=50$ and $m_0=u_0=100$ for DVSHGP, resulting in $n_0 = 200$ data points assigned to each expert; we have $m_{\mathrm{b}}=300$ basis functions for GPVC; we have $m=300$ for SVSHGP, FITC and VSGP; we have $m=300$ and $M=50$ for dPIC; we have $M=50$ for RBCM; finally, we train two separate GPs on a subset of size $m_{\mathrm{sod}}=2,000$ for EHSoD. As for optimization, DVSHGP adopts a two-stage process: it first only optimizes the variational parameters using CGD with up to 30 line searches, and then learns all the parameters jointly using up to 70 line searches; SVSHGP trains with NGD+Adam using $|\mathcal{B}|=1,000$ over 1,000 iterations; VSGP, FITC, GPVC and RBCM use up to 100 line searches to learn the parameters; dPIC employs the default optimization settings in the published codes; and finally EHSoD uses up to 50 line searches to train the two standard GPs, respectively.

\begin{figure}[!htb] 
	\centering
	\includegraphics[width=3.2in]{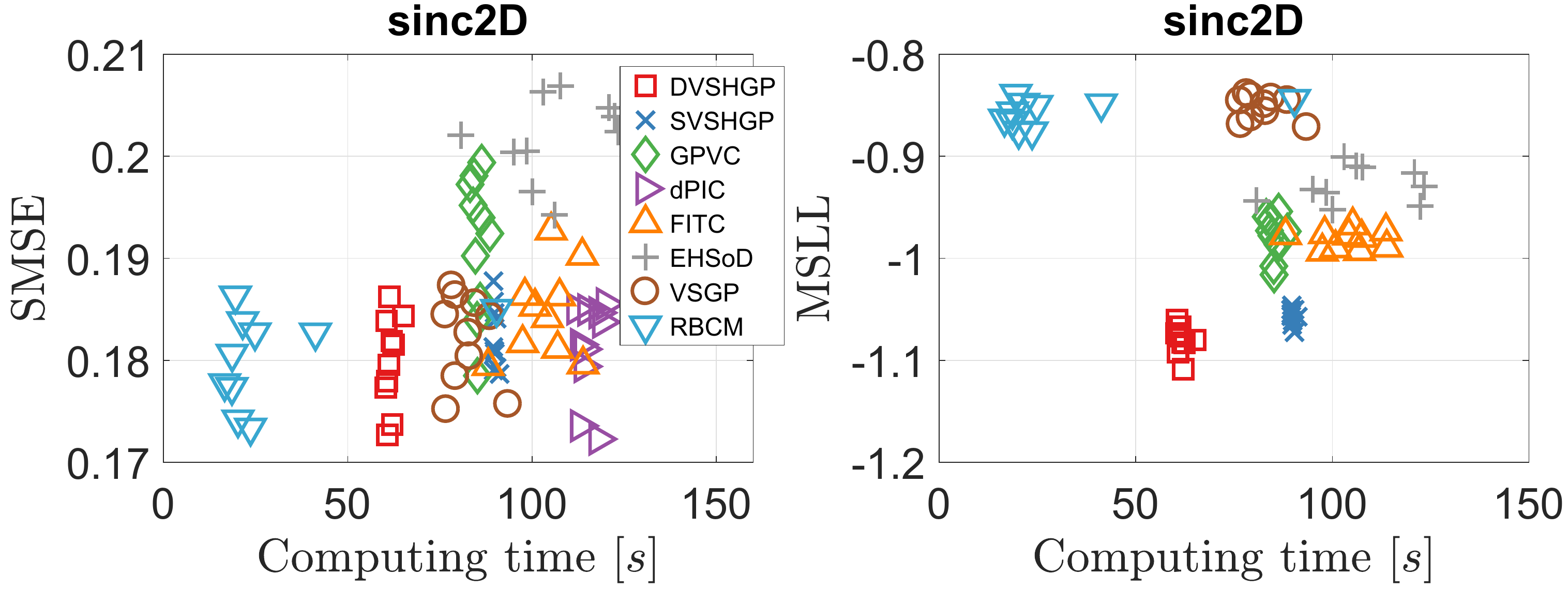}
	\caption{Comparison of GPs over ten runs on the synthetic dataset.}
	\label{Fig_sinc2D_smse_msll} 
\end{figure}

Fig.~\ref{Fig_sinc2D_smse_msll} depicts the modeling results of different GPs over ten runs on the synthetic \textit{sinc2D} dataset.\footnote{Since the dPIC codes only provide the prediction mean, we did not offer its MSLL value as well as the estimated noise variance in the following plots.} The horizontal axis represents the sum of training and predicting time for a model. It turns out that DVSHGP, SVSHGP, dPIC, VSGP and RBCM are competitive in terms of SMSE; but DVSHGP and SVSHGP perform better in terms of MSLL due to the well estimated heteroscedastic noise. Compared to the homoscedastic VSGP and RBCM, the FITC has heteroscedastic variances, which is indicated by the lower MSLL, at the cost of however (i) sacrificing the accuracy of prediction mean, and (ii) suffering from invalid noise variance $\sigma^2_{\epsilon}$.\footnote{FITC estimates $\sigma^2_{\epsilon}$ as 0.0030, while VSGP estimates it as 0.0309.} As a principled HGP, GPVC performs slightly better than FITC in terms of MSLL. Finally, the simple EHSoD has the worst SMSE; but it outperforms the homoscedastic VSGP and RBCM in terms of MSLL.

\begin{figure*}[!t] 
	\centering
	\includegraphics[width=6.0in]{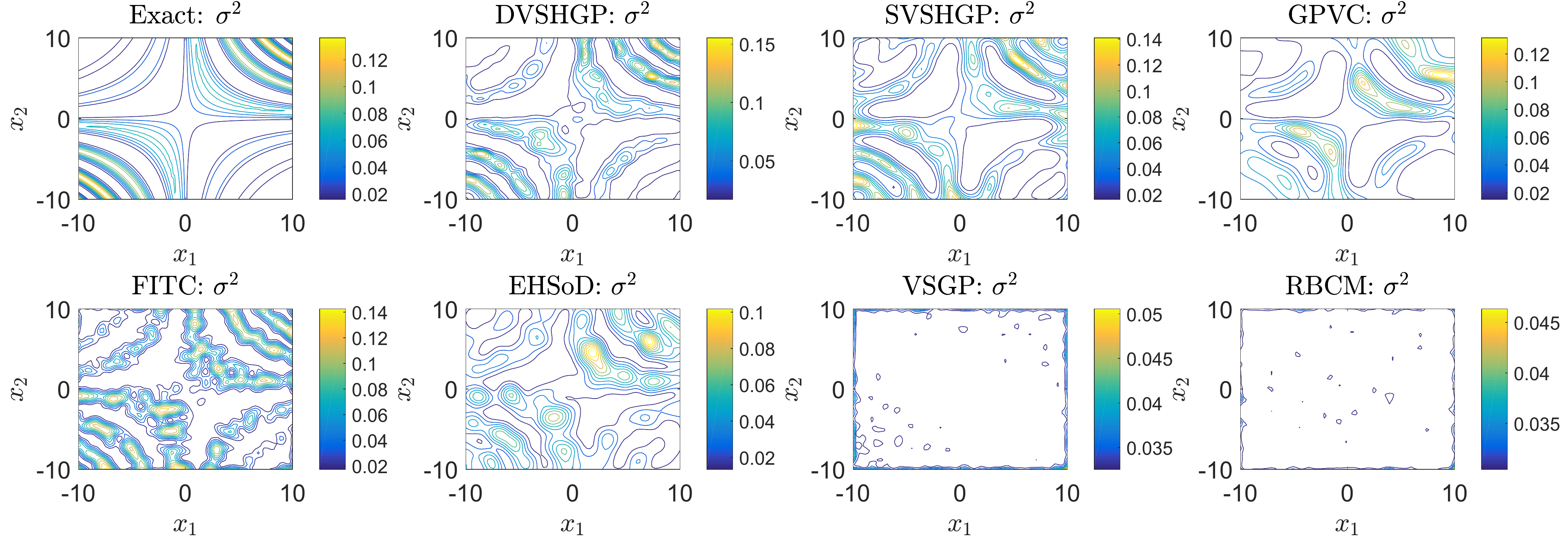}
	\caption{The exact noise variance and the prediction variances of different heteroscedastic/homoscedastic GPs on the synthetic dataset.}
	\label{Fig_sinc2D_GPs} 
\end{figure*}

In terms of efficiency, the RBCM requires less computing time since it contains no variational/inducing parameters, thus resulting in (i) lower complexity, and (ii) early stop for optimization. This also happens for the three datasets below.

Finally, Fig.~\ref{Fig_sinc2D_GPs} depicts the prediction variances of all the GPs except dPIC in comparison to the exact $\sigma^2$ on the synthetic dataset. It is first observed that the homoscedastic VSGP and RBCM are unable to describe the complex noise variance: they yield a nearly constant variance over the input domain. In contrast, DVSHGP, SVSHGP and GPVC capture the varying noise variance accurately by using an additional noise process $g$; FITC also captures the profile of the exact $\sigma^2$ with however unstable peaks and valleys; EHSoD is found to capture a rough expression of the exact $\sigma^2$.

\subsection{Medium real-world datasets}
This section conducts comparison on three real-world datasets. The first is the 9D protein dataset \cite{Dua:2017} with 45,730 data points. This dataset, taken from CASP 5-9, describes the physicochemical properties of protein tertiary structure. The second is the 21D sarcos dataset~\cite{rasmussen2006gaussian}, which relates to the inverse kinematics of a robot arm, has 48,933 data points. The third is the 3D \textit{3droad} dataset which comprises 434,874 data points \cite{kaul2013building} extracted from a 2D road network in North Jutland, Denmark, plussing elevation information. 

\subsubsection{The protein dataset}
For the \textit{protein} dataset, we randomly choose 35,000 training points and 10,730 test points. In the comparison, we have $M = 100$ (i.e., $n_0 = 350$) and $m_0 = u_0 =175$ for DVSHGP; we have $m=400$ for SVSHGP, VSGP and FITC; we have $m=400$ and $M=100$ for dPIC; we have $m_{\mathrm{b}}=400$ for GPVC; we have $M=100$ for RBCM; and finally we have $m_{\mathrm{sod}}=4,000$ for EHSoD. As for optimization, SVSHGP trains with NGD+Adam using $|\mathcal{B}|=2,000$ over 2,000 iterations. The optimization settings of other GPs keep consistent to that for the synthetic dataset. 

\begin{figure}[!t] 
	\centering
	\includegraphics[width=3.5in]{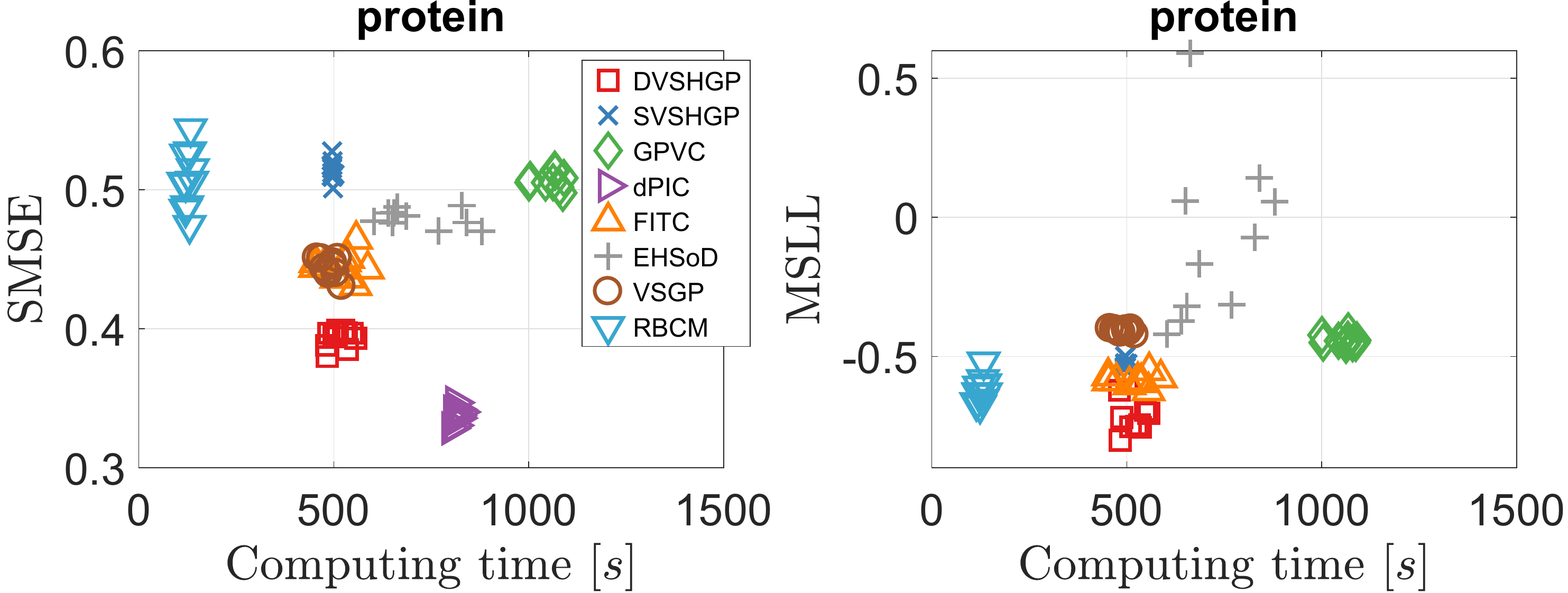}
	\caption{Comparison of GPs over ten runs on the \textit{protein} dataset.}
	\label{Fig_protein_smse_msll} 
\end{figure}

The results of different GPs over ten runs are summarized in Fig.~\ref{Fig_protein_smse_msll}. Among the HGPs, it is observed that dPIC outperforms the others in terms of SMSE, followed by DVSHGP. On the other hand, DVSHGP performs the best in terms of MSLL, followed by FITC and SVSHGP. The simple EHSoD is found to produce unstable MSLL results because of the small subset. Finally, the homoscedastic VSGP and RBCM provide mediocre SMSE and MSLL results.

\begin{figure}[!t] 
	\centering
	\includegraphics[width=3.0in]{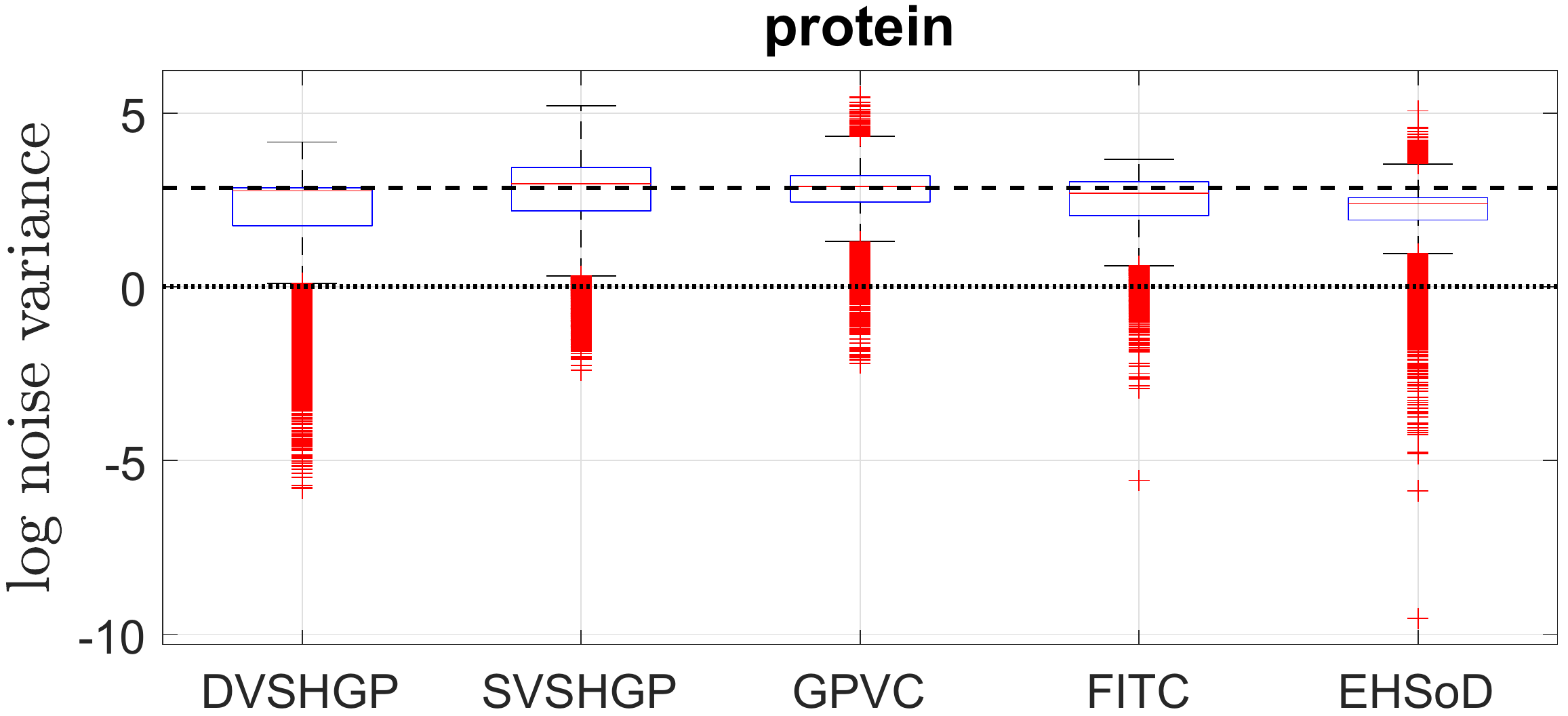}
	\caption{The distributions of log noise variances estimated by different GPs on the \textit{protein} dataset. The dash and dot lines indicate the log noise variances of VSGP and RBCM, respectively.}
	\label{Fig_protein_noise_var} 
\end{figure}

Next, Fig.~\ref{Fig_protein_noise_var} offers insights into the distributions of log noise variances of all the GPs except dPIC on the \textit{protein} dataset for a single run. Note that (i) as homoscedastic GPs, the log noise variances of VSGP and RBCM are marked as dash and dot lines, respectively; and (ii) we plot the variance of $p(f_*|\mathcal{D}, \bm{x}_*)$ for FITC since (a) it accounts for the heteroscedasticity and (b) the scalar noise variance $\sigma^2_{\epsilon}$ is severely underestimated. The results in Fig.~\ref{Fig_protein_noise_var} indicate that the \textit{protein} dataset may contain heteroscedastic noise. Besides, compared to the VSGP which uses a global inducing set, the localized RBCM provides a more compact estimation of $\sigma^2_{\epsilon}$. This compact noise variance, which has also been observed on the two datasets below, brings lower MSLL for RBCM.

Furthermore, we clearly observe the interaction between $f$ and $g$ for DVSHGP, SVSHGP and GPVC. The small MSLL of RBCM suggests that the \textit{protein} dataset may own small noise varaicnes at some test points. Hence, the localized DVSHGP, which is enabled to capture the local variety through \textit{individual} variational and inducing parameters for each expert, produces a longer tail in Fig.~\ref{Fig_protein_noise_var}. The well estimated heteroscedastic noise in turn improves the prediction mean of DVSHGP through the interaction between $f$ and $g$. In contrast, due to the limited global inducing set, the prediction mean of SVSHGP and GPVC is traded for capturing heteroscedastic noise.

\begin{figure}[!t] 
	\centering
	\includegraphics[width=3.5in]{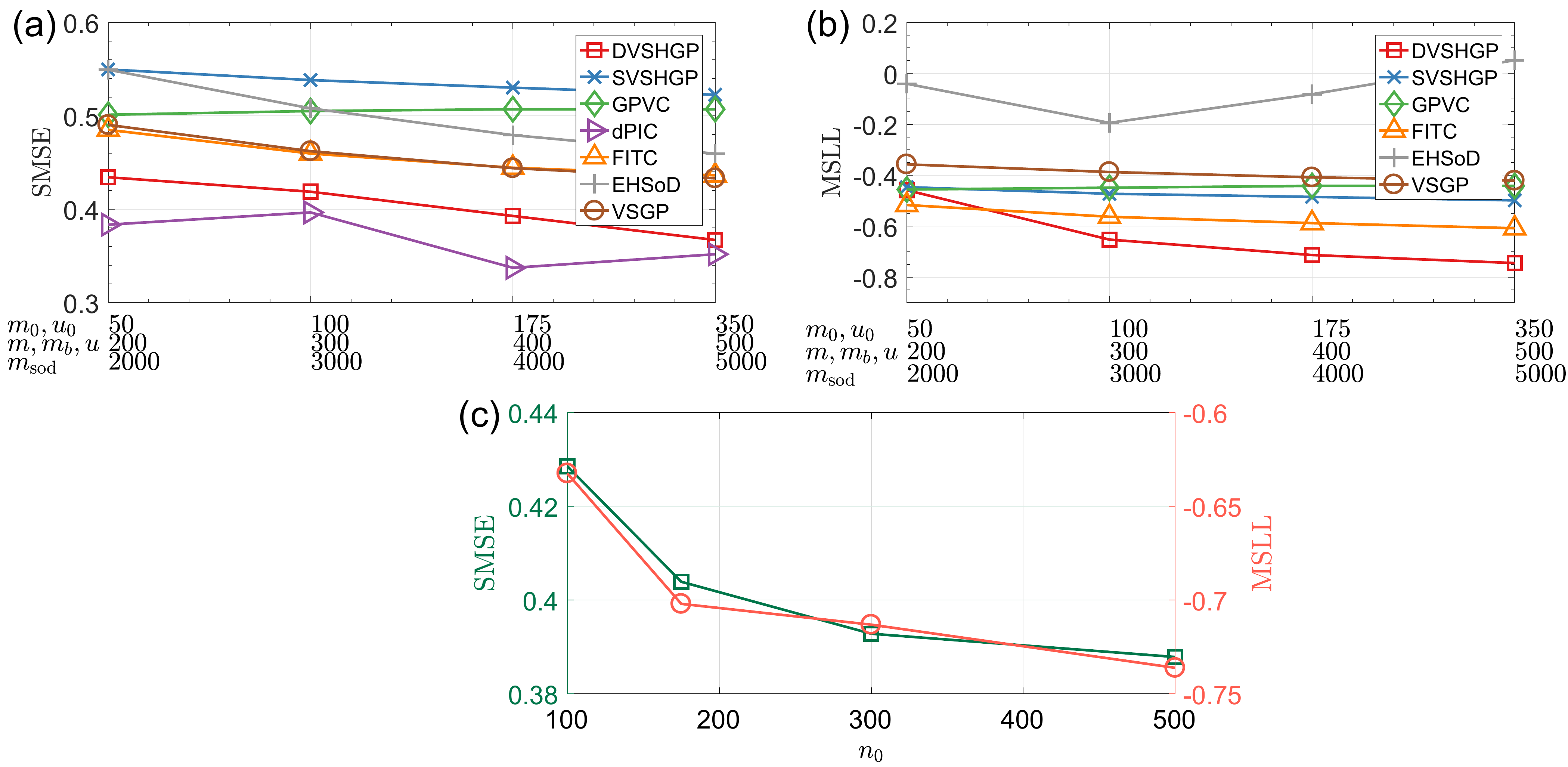}
	\caption{The effect of algorithmic parameters on the performance of different sparse GPs on the \textit{protein} dataset.}
	\label{Fig_protein_varyParas} 
\end{figure}

Notably, the performance of sparse GPs is affected by their modeling parameters, e.g., the inducing sizes $m_0$, $m$, $u_0$ and $u$, the number of basis functions $m_{\mathrm{b}}$, and the subset size $m_{\mathrm{sod}}$. Fig.~\ref{Fig_protein_varyParas}(a) and (b) depict the average results of sparse GPs over ten runs using different parameters. Particularly, we  investigate the impact of subset size $n_0$ on DVSHGP in Fig.~\ref{Fig_protein_varyParas}(c) using $m_0=u_0=0.5n_0$. It is found that DVSHGP favours large $n_0$ (small $M$) and large $m_0$ and $u_0$. Similarly, VSGP and FITC favour more inducing points. However, dPIC offers an unstable SMSE performance with increasing $m$; GPVC performs slightly worse with increasing $m_{\mathrm{b}}$ in terms of both SMSE and MSLL, which has also been observed in the original paper \cite{almosallam2016gpz}, and may be caused by the sharing of basis functions for $f$ and $g$. Finally, EHSoD showcases poorer MSLL values when $m_{\mathrm{sod}} \ge 3000$, because of the difficulty of approximating the empirical variances.

\subsubsection{The sarcos dataset}
For the \textit{sarcos} dataset, we randomly choose 40,000 training points and 8,933 test points. In the comparison, we have $M = 120$ (i.e., $n_0 \approx 333$) and $m_0=u_0=175$ for DVSHGP; we have $m=600$ for SVSHGP, VSGP and FITC; we have $m=600$ and $M=120$ for dPIC; we have $m_{\mathrm{b}}=600$ for GPVC; we have $M=120$ for RBCM; and finally we have $m_{\mathrm{sod}}=4,000$ for EHSoD. The optimization settings are the same as before. 

\begin{figure}[!t] 
	\centering
	\includegraphics[width=3.3in]{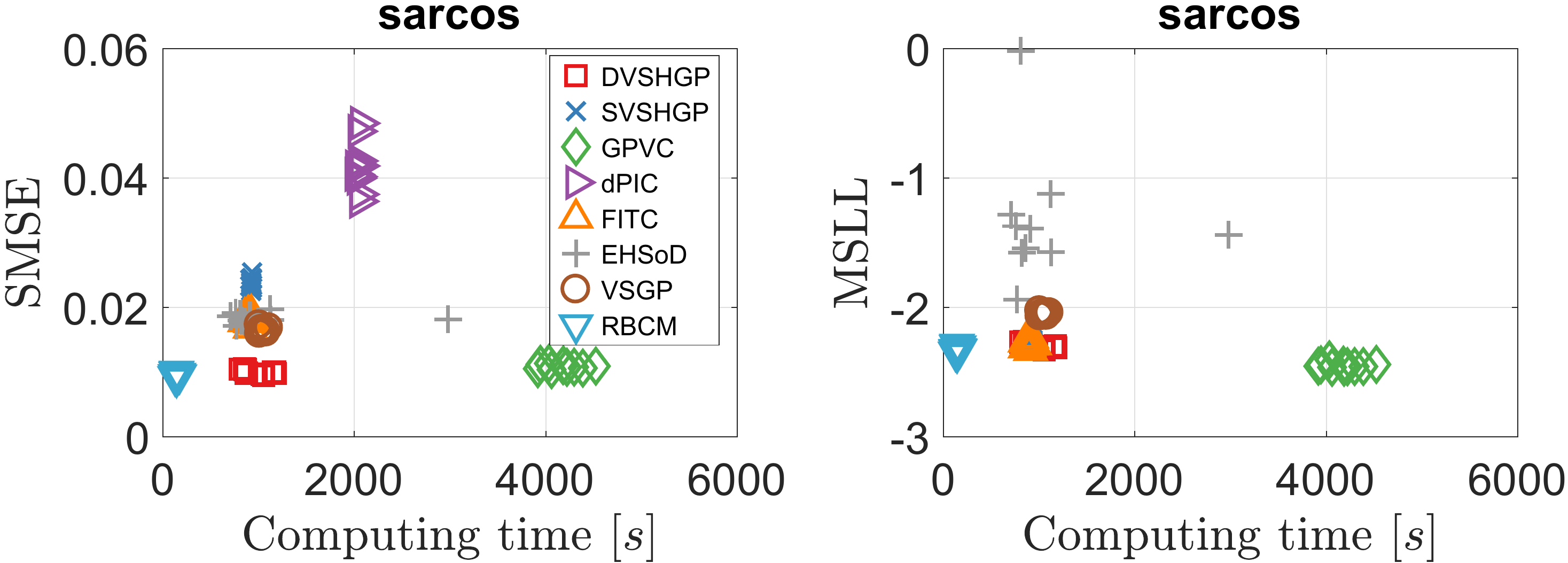}
	\caption{Comparison of GPs over ten runs on the \textit{sarcos} dataset.}
	\label{Fig_sarcos_smse_msll} 
\end{figure}

\begin{figure}[!t] 
	\centering
	\includegraphics[width=3.0in]{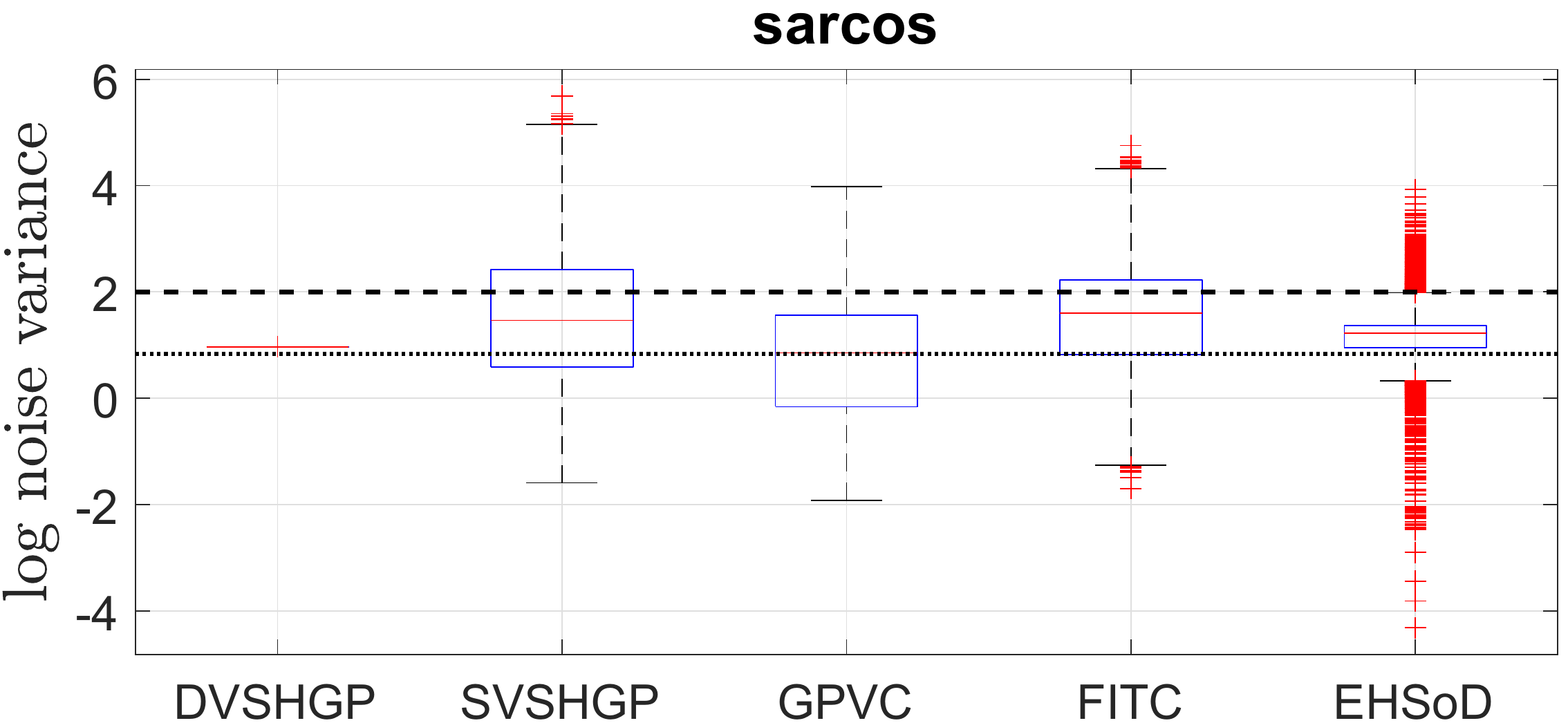}
	\caption{The distributions of log noise variances of different GPs on the \textit{sarcos} dataset. The dash and dot lines indicate the log noise variances of VSGP and RBCM, respectively.}
	\label{Fig_sarcos_y_mu} 
\end{figure}

The results of different GPs over ten runs on the \textit{sarcos} dataset are depicted in Fig.~\ref{Fig_sarcos_smse_msll}. Besides, Fig.~\ref{Fig_sarcos_y_mu} depicts the log noise variances of the GPs on this dataset. Different from the \textit{protein} dataset, the \textit{sarcos} dataset seems to have weak heteroscedastic noises across the input domain, which is verified by the facts that (i) the noise variance of DVSHGP is a constant, and (ii) DVSHGP agrees with RBCM in terms of both SMSE and MSLL. Hence, all the HGPs except EHSoD perform similarly in terms of MSLL. 

In addition, the weak heteroscedasticity in the \textit{sarcos} dataset reveals that we can use only a few inducing points for $\bm{g}$ to speed up the inference. For instance, we retrain DVSHGP using $u_0 = 5$. This extremely small inducing set for $\bm{g}$ brings (i) much less computing time of about 350 seconds, and (ii) almost the same model accuracy with SMSE = 0.0099 and MSLL = -2.3034.

\subsubsection{The 3droad dataset}
Finally, for the \textit{3droad} dataset, we randomly choose 390,000 training points, and use the remaining 44,874 data points for testing.  In the comparison, we have $M = 800$ (i.e., $n_0 \approx 487$) and $m_0=u_0=250$ for DVSHGP; we have $m=500$ for SVSHGP, VSGP and FITC; we have $m=500$ and $M=800$ for dPIC; we have $m_{\mathrm{b}}=500$ for GPVC; we have $M=800$ for RBCM; and finally we have $m_{\mathrm{sod}}=8,000$ for EHSoD. As for optimization, SVSHGP trains with NGD+Adam using $|\mathcal{B}|=4,000$ over 4,000 iterations. The optimization settings of other GPs keep the same as before. 

\begin{figure}[!t] 
	\centering
	\includegraphics[width=3.2in]{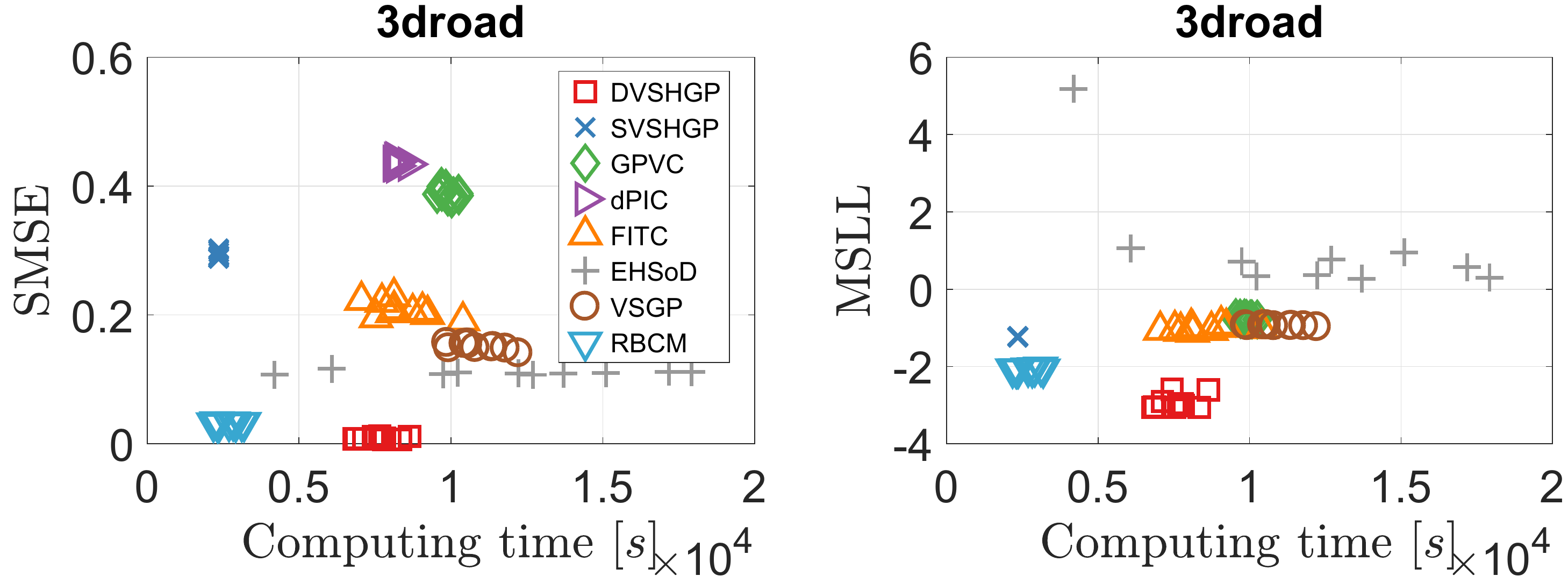}
	\caption{Comparison of GPs over ten runs on the \textit{3droad} dataset.}
	\label{Fig_3droad_smse_msll} 
\end{figure}

The results of different GPs over ten runs on the \textit{3droad} dataset are depicted in Fig.~\ref{Fig_3droad_smse_msll}. It is observed that DVSHGP outperforms the others in terms of both SMSE and MSLL, followed by RBCM. For other HGPs, especially SVSHGP and GPVC, the relatively poor noise variance (large MSLL) in turn sacrifices the accuracy of prediction mean. Even though, the heteroscedastic noise helps SVSHGP, GPVC and FITC perform similarly to VSGP in terms of MSLL. 

\begin{figure}[!t] 
	\centering
	\includegraphics[width=3.0in]{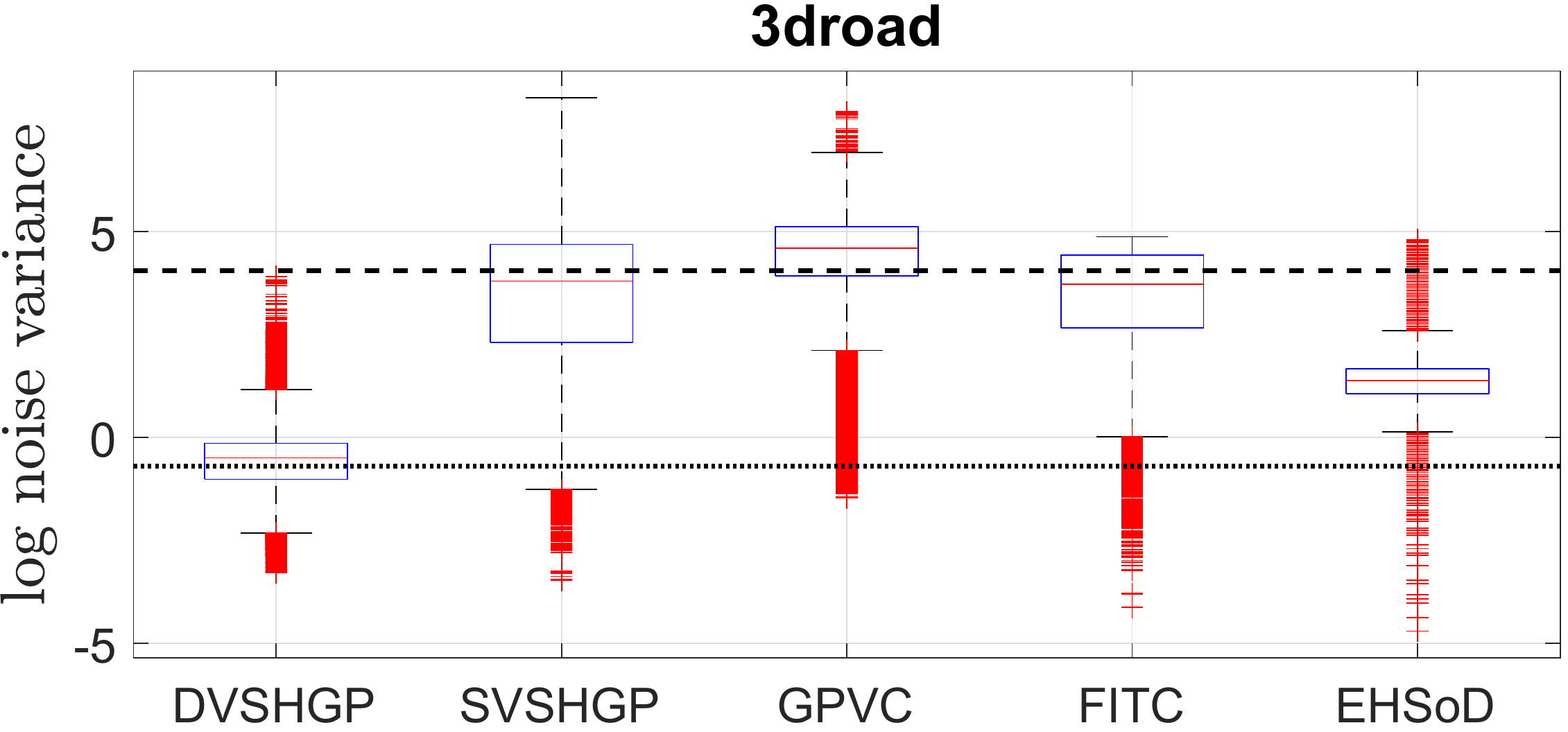}
	\caption{The distributions of log noise variances of different GPs on the \textit{3droad} dataset. The dash and dot lines indicate the log noise variances of VSGP and RBCM, respectively.}
	\label{Fig_3droad_y_mu} 
\end{figure}

In addition, Fig.~\ref{Fig_3droad_y_mu} depicts the log noise variances of these GPs on the \textit{3droad} dataset. The highly accurate prediction mean of DVSHGP helps well estimate the heteroscedastic noise. It is observed that (i) the noise variances estimated by DVSHGP are more compact than that of other HGPs; and (ii) the average noise variance agrees with that of RBCM.

Finally, the results from the \textit{3droad} dataset together with the other two datasets indicate that:
\begin{itemize}
	\item the well estimated noise variance of HGPs in turn improves the prediction mean via the interaction between $f$ and $g$; otherwise, it may sacrifice the prediction mean;
	\item the heteroscedastic noise usually improves sparse HGPs over the homoscedastic VSGP in terms of MSLL.
\end{itemize}

\subsection{Large real-world dataset}
The final section evaluates the performance of different GPs on the 11D \textit{electric} dataset,\footnote{The dataset is available at \url{https://archive.ics.uci.edu/ml/index.php}.} which is partitioned into two million training points
and 49,280 test points. The HGPs in the comparison include DVSHGP, SVSHGP, dPIC and EHSoD.\footnote{GPVC and FITC are unaffordable for this massive dataset. Besides, the stochastic variant of FITC~\cite{hoang2015unifying} is not included, since it does not support end-to-end training.} Besides, the RBCM and the stochastic variant of VSGP, named SVGP~\cite{hensman2013gaussian}, are employed for this big dataset.

\begin{table}[!t]
	\centering
	\caption{Comparison of GPs on the \textit{electric} dataset.}
	\label{Tab_large_realworld_datasets}
	\resizebox{\columnwidth}{!}{%
	\begin{tabular}{|c|c|c|c|c|c|c|}		
		\hline
		& DVSHGP & SVSHGP & dPIC &
		EHSoD & SVGP & RBCM \\
		\hline
		\hline
		SMSE & 0.0020 & 0.0029 & 0.0042 &
		0.0103 & 0.0028 & 0.0023 \\
		\hline
		MSLL & -3.4456 & -3.1207 & - &
		-1.9453 & -2.8489 & -3.0647 \\
		\hline
		$t$ [$h$] & 11.05 & 7.44 & 47.22 &
		4.72 & 3.55 & 3.97 \\
		\hline
	\end{tabular}
    }
\end{table}

In the comparison, we have $M = 2,000$ (i.e., $n_0 = 1,000$) and $m_0=u_0=300$ for DVSHGP; we have $m=2,500$ for SVSHGP and SVGP; we have $m=2,500$ and $M=2,000$ for dPIC; we have $M=2,000$ for RBCM; and finally we have $m_{\mathrm{sod}}=15,000$ for EHSoD. As for optimization, SVSHGP trains with NGD+Adam using $|\mathcal{B}|=5,000$ over 10,000 iterations; The optimization settings of other GPs keep the same as before. 

The average results over five runs in Table~\ref{Tab_large_realworld_datasets} indicate that DVSHGP outperforms the others in terms of both SMSE and MSLL, followed by SVSHGP. The simple EHSoD provides the worst performance, and cannot be improved by using larger $m_{\mathrm{sod}}$ due to the memory limit in current infrastructure. Additionally, in terms of efficiency, we find that (i) SVSHGP is better than DVSHGP due to the parallel/GPU acceleration deployed in Tensorflow;\footnote{Further GPU speedup could be utilized for DVSHGP in Matlab.} (ii) SVGP is better than SVSHGP because of lower complexity; and (iii) the huge computing time of dPIC might be incurred by the unoptimized codes. 

\begin{figure}[!t] 
	\centering
	\includegraphics[width=3.2in]{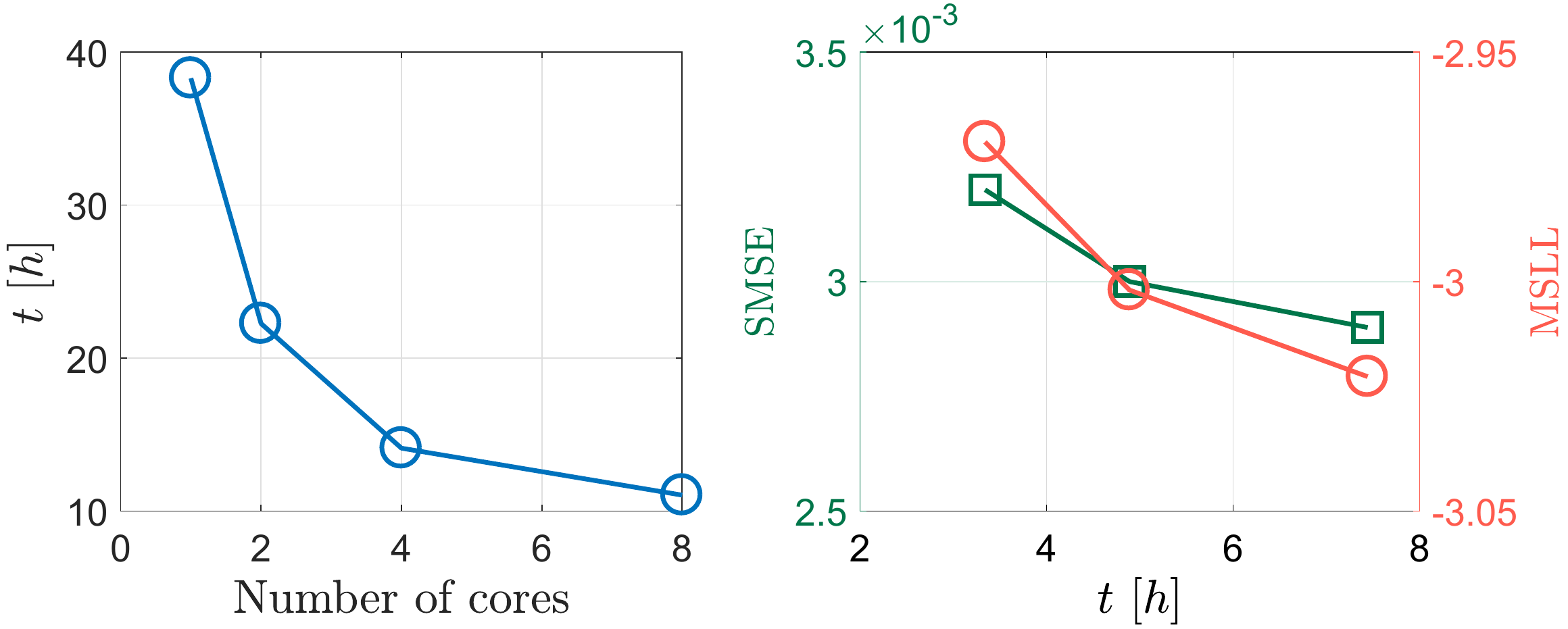}
	\caption{Illustration of (a) the computing time of DVSHGP vs. number of parallel cores, and (b) the performance of SVSHGP, from left to right, using $|\mathcal{B}|=1,000$, 2,500 and 5,000 on the \textit{electric} dataset.}
	\label{Fig_elec_num_cores} 
\end{figure}

Finally, due to the distributed framework, Fig.~\ref{Fig_elec_num_cores}(a) depicts the total computing time of DVSHGP using different numbers of processing cores. It is observed that the DVSHGP using eight cores achieves a speedup around 3.5 in comparison to the centralized counterpart. Fig.~\ref{Fig_elec_num_cores}(b) also exploits the performance of SVSHGP using a varying mini-batch size $|\mathcal{B}|$. It is observed that (i) a small $|\mathcal{B}|$ significantly speeds up the model training, and (ii) different mini-batch sizes yield similar SMSE and MSLL here, because the model has been optimized over sufficient iterations.

\section{Discussions and conclusions} \label{sec_conclusion}
In order to scale up the original HGP~\cite{goldberg1998regression}, we have presented distributed and stochastic variational sparse HGPs. The proposed SVSHGP improves the scalability through stochastic variational inference. The proposed DVSHGP (i) enables large-scale regression via distributed computations, and (ii) achieves high model capability via localized experts and many inducing points. We compared them to existing scalable homoscedastic/heteroscedastic GPs on a synthetic dataset and four real-world datasets. The comparative results obtained indicated that DVSHGP exhibits superior performance in terms of both SMSE and MSLL; while due to the limited global inducing set, SVSHGP may sacrifice the prediction mean for capturing heteroscedastic noise.

Apart from our scalable HGP framework, there are some new GP paradigms developed recently from different perspectives for improving the predictive distribution. For instance, instead of directly modeling the heteroscedastic noise, we could introduce additional latent inputs to modulate the kernel~\cite{wang2012gaussian, dutordoir2018gaussian}; or we directly target the interested posterior distribution to enhance the prediction of $f$~\cite{jankowiak2019sparse}; or we adopt highly flexible priors, e.g., implicit processes, over functions~\cite{ma2019variational}; or we mix various GPs during the inference~\cite{nguyen2014fast,wu2018two}; or we develop the specific non-stationary kernel~\cite{remes2017non}. They bring new interpretations at the cost of however losing the direct description of heteroscedasticity or raising complicated inference with high complexity. But all these paradigms together with our scalable HGPs greatly enable future exploitation of fitting the data distribution with high quality and efficiency.

Finally, our future work will consider the heteroscedasticity in the underlying function $f$, i.e., the non-stationary, such as \cite{heinonen2016non, munoz2016laplace, nguyen2014fast}. The integration of various kinds of heteroscedasticity is believed to improve predictions.


%

\appendices
\section{Non-negativity of $\Lambda_{nn}$} \label{APP_Non_negativity}
We know that the variational diagonal matrix $\bm{\Lambda}_{nn}$ expresses
\begin{equation*}
\bm{\Lambda}_{nn} = 0.5 (\bm{\Lambda}_{nn}^a + \bm{\Lambda}_{nn}^b + \bm{I}).
\end{equation*}
In order to prove the non-negativity of $\bm{\Lambda}_{nn}$, we should figure out the non-negativity of the diagonal elements of $\bm{\Lambda}^a_{nn} + \bm{I}$ and $\bm{\Lambda}^b_{nn}$, respectively.

Firstly, the diagonal elements of $\bm{\Lambda}^b_{nn}$ write
\begin{equation*}
[\bm{\Lambda}_{nn}^b]_{ii} = [(\bm{K}_{nn}^f - \bm{Q}_{nn}^f) \bm{R}_g^{-1}]_{ii}, \quad 1 \le i \le n,
\end{equation*}
where the diagonal elements of $\bm{R}_g^{-1}$ satisfy $[\bm{R}_g^{-1}]_{ii} = e^{[\bm{\Sigma}_{g}]_{ii}/2 - [\bm{\mu}_{g}]_i} > 0$, $1 \le i \le n$; and the diagonal elements of $\bm{K}_{nn}^f - \bm{Q}_{nn}^f$ are the variances of training conditional $p(\bm{f} | \bm{f}_m)$. Therefore, $\bm{\Lambda}_{nn}^b$ has non-negative diagonal elements.

Secondly, the diagonal elements of $\bm{\Lambda}^a_{nn} + \bm{I}$ write, given $\bm{\beta}_n =\bm{\Sigma}_y^{-1} \bm{y}$,
\begin{equation*}
[\bm{\Lambda}_{nn}^a + \bm{I}]_{ii} = [\bm{\beta}_n \bm{\beta}_n^{\mathsf{T}} \bm{R}_g - \bm{\Sigma}_y^{-1} \bm{R}_g  + \bm{I}]_{ii}, \quad 1 \le i \le n.
\end{equation*}
For $\bm{\beta}_n \bm{\beta}_n^{\mathsf{T}} \bm{R}_g$, the diagonal elements are non-negative. For $\bm{I} - \bm{\Sigma}_y^{-1} \bm{R}_g$, given the Cholesky decomposition $\bm{K}_{\Lambda}^f = \bm{L}_{\Lambda}^f (\bm{L}_{\Lambda}^f)^{\mathsf{T}}$, we have
\begin{equation*}
\begin{aligned}
\bm{I} - \bm{\Sigma}_y^{-1} \bm{R}_g =& [\bm{R}_g^{-1} \bm{K}_{nm}^f (\bm{K}_{\Lambda}^f)^{-1} \bm{K}_{mn}^f \bm{R}_g^{-1}] \bm{R}_g \\
=& [\bm{R}_g^{-1} \bm{K}_{nm}^f (\bm{L}_{\Lambda}^f)^{-\mathsf{T}} (\bm{L}_{\Lambda}^f)^{-1} \bm{K}_{mn}^f \bm{R}_g^{-1}] \bm{R}_g,
\end{aligned}
\end{equation*}
indicating that the diagonal elements must be non-negative.

Hence, from the foregoing discussions, we know that $\bm{\Lambda}_{nn}$ is a non-negative diagonal matrix.

\section{Derivatives of $F_V$ w.r.t. hyperparameters} \label{APP_Derivs_F}
Let $\bm{\lambda}_n = \log (\bm{\Lambda}_{nn} \bm{1})$ collect $n$ variational parameters in the log form for non-negativity, we have the derivatives of $F_V$ w.r.t. $\bm{\lambda}_{n}$ as
\begin{equation*}
\begin{aligned}
\frac{\partial F_V}{\partial \bm{\lambda}_{n}} =& \bm{\Lambda}_{nn} \left[\frac{1}{2} (\bm{Q}_{nn}^g + \frac{1}{2} \bm{A}_{nn}) \bm{\Lambda}_{nn}^{ab} \bm{1} + \frac{1}{4} \bm{A}_{nn} \bm{1} \right. \\
&-\left.  \frac{1}{2}\bm{A}_{nn} \bm{\Lambda}_{nn} \bm{1} - \bm{\mu}_g + \mu_0\bm{1} \right],
\end{aligned}
\end{equation*}
where $\bm{A}_{nn} = (\bm{K}_{nu}^g \bm{K}_{\Lambda}^{-1} \bm{K}_{un}^g)^{\odot 2}$, and the operator ${\odot 2}$ represents the element-wise power.

The derivatives of $F_V$ w.r.t. the kernel parameters $\bm{\theta}^f = \{\theta_i^f\}$ are
\begin{equation*}
\frac{\partial F_V}{\partial \theta^f_i} = \frac{1}{2} \mathrm{Tr} \left[ (\bm{\beta}_n \bm{\beta}_n^{\mathsf{T}} - \bm{\Sigma}_y^{-1} + \bm{R}_g^{-1}) \frac{\partial \bm{Q}_{nn}^f}{\partial \theta^f_i} - \bm{R}_g^{-1} \frac{\partial \bm{K}_{nn}^f}{\partial \theta^f_i}\right].
\end{equation*}
Similarly, the derivatives of $F_V$ w.r.t. the kernel parameters $\bm{\theta}^g = \{\theta_i^g\}$ are
\begin{equation*}
\begin{aligned}
\frac{\partial F_V}{\partial \theta^g_i} =& \frac{1}{2} \mathrm{Tr} \left[\frac{\partial \bm{\mu}_{g}}{\partial \theta^g_i} (\bm{1}^{\mathsf{T}} \bm{\Lambda}_{nn}^{ab}) - \frac{1}{2} (\bm{\Lambda}_{nn}^{ab} + \bm{I}) \frac{\partial \bm{\Sigma}_{g}}{\partial \theta^g_i} \right] \\
-& \frac{1}{2}\mathrm{Tr}\left[ \bm{V}_{uu}^g \frac{\partial \bm{K}_{uu}^g }{\partial \theta^g_i} - (\bm{\Omega}_{nu}^g)^{\mathsf{T}} \bm{\Lambda}_{nn} \bm{\Omega}_{nu}^g \frac{\partial \bm{\Sigma}_u}{\partial \theta^g_i} + 2\frac{\partial \bm{\mu}_u}{\partial \theta^g_i} \bm{\gamma}_{u}^{\mathsf{T}} \right],
\end{aligned}
\end{equation*}
where $\bm{\gamma}_{u} = (\bm{K}_{uu}^g)^{-1} (\bm{\mu}_{u} -\mu_0\bm{1})$ and $\bm{V}_{uu}^g = (\bm{K}_{uu}^g)^{-1} - \bm{\gamma}_{u}\bm{\gamma}_{u}^{\mathsf{T}} - (\bm{K}_{uu}^g)^{-1} \bm{\Sigma}_{u} (\bm{K}_{uu}^g)^{-1}$.

The derivatives of $F_V$ w.r.t. the mean parameter $\mu_0$ of $g$ is
\begin{equation*}
\frac{\partial F_V}{\partial \mu_0} = \frac{1}{2} \mathrm{Tr} (\bm{\Lambda}_{nn}^{ab}).
\end{equation*}

Finally, we calculate the derivatives of $F_V$ w.r.t. the inducing points $\bm{X}_m$ and $\bm{X}_u$. Since the inducing points are involved in the kernel matrices, we get the derivatives $\partial \bm{K}^f_{nm} / \partial x^f_{ij}$, $\partial \bm{K}^f_{mn} / \partial x^f_{ij}$, $\partial \bm{K}^f_{mm} / \partial x^f_{ij}$, $\partial \bm{K}^g_{nu} / \partial x^g_{ij}$, $\partial \bm{K}^g_{un} / \partial x^g_{ij}$, and $\partial \bm{K}^g_{uu} / \partial x^g_{ij}$, where $x^f_{ij} = [\bm{X}_m]_{ij}$ and $x^g_{ij} = [\bm{X}_u]_{ij}$. We first obtain the derivatives of $F_V$ w.r.t. $\bm{X}_m$ as
\begin{equation*}
\frac{\partial F_V}{\partial x^f_{ij}} = 2\mathrm{Tr} \left[ \frac{\partial \bm{K}^f_{nm}}{\partial x^f_{ij}} \bm{A}^f_{mn} \right] + \mathrm{Tr} \left[ \frac{\partial \bm{K}^f_{mm}}{\partial x^f_{ij}} \bm{A}^f_{mm} \right],
\end{equation*}
where $\bm{A}_{mn}^f = 0.5 (\bm{\Omega}_{nm}^f)^{\mathsf{T}} (\bm{\beta}_n \bm{\beta}_n^{\mathsf{T}} - \bm{\Sigma}_y^{-1} + \bm{R}_g^{-1})$, and $\bm{A}_{mm}^f = -\bm{A}_{mn}^f \bm{\Omega}_{nm}^f$.
Similarly, the derivatives of $F_V$ w.r.t. $\bm{X}_u$ write
\begin{equation*}
\frac{\partial F_V}{\partial x^g_{ij}} = \mathrm{Tr} \left[ \frac{\partial \bm{K}^g_{nu}}{\partial x^g_{ij}} \bm{A}^g_{un} \right] + \mathrm{Tr} \left[ \bm{A}^g_{nu} \frac{\partial \bm{K}^g_{un}}{\partial x^g_{ij}} \right] + \mathrm{Tr} \left[ \frac{\partial \bm{K}^g_{uu}}{\partial x^g_{ij}} \bm{A}^g_{uu} \right].
\end{equation*}
For $\bm{A}_{un}^g$ in $\partial F_V / \partial x^g_{ij}$, we have
\begin{equation*}
\begin{aligned}
\bm{A}^g_{un} =& \underbrace{0.5 (\bm{\Omega}_{nu}^g)^{\mathsf{T}} (\bm{\Lambda}_{nn} - 0.5 \bm{I}) \bm{1} (\bm{1}^{\mathsf{T}} \bm{\Lambda}_{nn}^{ab})}_{\bm{T}_1} \\
&+ \underbrace{0.25 \left( \bm{H}_{uu} \bm{K}_{un}^g  \bm{\Lambda}_{nn} - [\bm{\Omega}_{nu}^{\Lambda}+ \bm{\Omega}_{nu}^g]^{\mathsf{T}}  (\bm{\Lambda}_{nn}^{ab} + \bm{I}) \right)}_{\bm{T}_2} \\
&+ \underbrace{0.5 \left( \bm{J}_{uu} \bm{K}_{un}^g \bm{\Lambda}_{nn} - \bm{\gamma}_{u} \bm{1}^{\mathsf{T}} (\bm{\Lambda}_{nn} - 0.5 \bm{I}) \right)}_{\bm{T}_3}.
\end{aligned}
\end{equation*}
where $\bm{H}_{uu} = (\bm{\Omega}_{nu}^{\Lambda})^{\mathsf{T}} (\bm{\Lambda}_{nn}^{ab} + \bm{I}) \bm{\Omega}_{nu}^{\Lambda}$, and $\bm{J}_{uu} = \bm{K}_{\Lambda}^{-1} \bm{K}_{uu}^g ((\bm{K}_{uu}^g)^{-1} - \bm{\Sigma}_u^{-1}) \bm{K}_{uu}^g \bm{K}_{\Lambda}^{-1}$. For $\bm{A}_{nu}^g$, we have
\begin{equation*}
\bm{A}^g_{nu} =0.5 (\bm{\Lambda}_{nn} - 0.5 \bm{I}) \bm{1} (\bm{1}^{\mathsf{T}} \bm{\Lambda}_{nn}^{ab}) \bm{\Omega}_{nu}^g + \bm{T}_2^{\mathsf{T}} + \bm{T}_3^{\mathsf{T}}.
\end{equation*}
For $\bm{A}_{uu}^g$, we have
\begin{equation*}
\begin{aligned}
\bm{A}_{uu}^g =& -\bm{T}_1 \bm{\Omega}_{nu}^g -0.25 \left( (\bm{\Omega}_{nu}^g)^{\mathsf{T}} (\bm{\Lambda}_{nn}^{ab} + \bm{I}) \bm{\Omega}_{nu}^g - \bm{H}_{uu} \right) \\
&+ 0.5 \left( \bm{P}_{uu} + \bm{P}_{uu}^{\mathsf{T}} + \bm{V}_{uu}^g - \bm{J}_{uu} \right),
\end{aligned}
\end{equation*}
where $\bm{P}_{uu} = \bm{K}_{\Lambda}^{-1} \bm{K}_{uu}^g ((\bm{K}_{uu}^g)^{-1} - \bm{\Sigma}_u^{-1})$.

The calculation of $\partial F_V / \partial x^f_{ij}$ and $\partial F_V / \partial x^g_{ij}$ requires a loop over $m \times d$ and $u \times d$ parameters of the inducing points, which is quite slow for even moderate $m$, $u$ and $d$. Fortunately, we know that the derivative $\partial \bm{K} / \partial x_{ij}^{f(g)}$ only has $n$ or $m$ ($u$) non-zero elements. Due to the sparsity, $\partial \bm{K} / \partial x_{ij}^{f(g)}$ can be performed in vectorized operations such that the derivatives w.r.t. all the inducing points can be calculated along a specific dimension.

\section{Natural gradients of $q(\bm{f}_m)$ and $q(\bm{g}_u)$}
\label{sec_natural_gradients}
For exponential family distributions\footnote{The probability density function (PDF) of exponential family is $p(\bm{x}) = h(\bm{x}) e^{\bm{\theta}^{\mathsf{T}}\bm{t}(\bm{x})-A(\bm{\theta})}$, where $\bm{\theta}$ is natural parameters, $h(\bm{x})$ is underlying measure, $\bm{t}(\bm{x})$ is sufficient statistic, and $A(\bm{\theta})$ is log normalizer. Besides, the expectation parameters are defined as $\bm{\psi} = \mathbb{E}_{p(\bm{x})} [\bm{t}(\bm{x})]$.
} parameterized by \textit{natural} parameters $\bm{\theta}$, we update the parameters using natural gradients as
\begin{equation*}
\bm{\theta}_{(t+1)} = \bm{\theta}_{(t)} - \gamma_{(t)} \bm{G}_{\bm{\theta}_{(t)}}^{-1} \frac{\partial F}{\partial \bm{\theta}_{(t)}} = \bm{\theta}_{(t)} - \gamma_{(t)} \frac{\partial F}{\partial \bm{\psi}_{(t)}},
\end{equation*}
where $F$ is the objective function, and $\bm{G}_{\bm{\theta}} = \partial \bm{\psi}_{(t)} / \partial \bm{\theta}_{(t)}$ is the fisher information matrix with $\bm{\psi}$ being the \textit{expectation} parameters of exponential distributions.

For $q(\bm{g}_u) = \mathcal{N}(\bm{g}_u| \bm{\mu}_u, \bm{\Sigma}_u)$, its natural parameters are $\bm{\theta}$ which are partitioned into two components
\begin{equation*}
\bm{\theta}_1 = \bm{\Sigma}_u^{-1}\bm{\mu}_u, \quad \bm{\Theta}_2 = -\frac{1}{2}\bm{\Sigma}_u^{-1},
\end{equation*}
where $\bm{\theta}_1$ comprises the first $m$ elements of $\bm{\theta}$, and $\mathbf{\Theta}_2$ the remaining elements reshaped to a square matrix. Accordingly, the expectation parameters $\bm{\psi}$ are divided as
\begin{equation*}
\bm{\psi}_1 = \bm{\mu}_u, \quad \bm{\Psi}_2 = \bm{\mu}_u \bm{\mu}_u^{\mathsf{T}} + \bm{\Sigma}_u.
\end{equation*}
Thereafter, we update the natural parameters with step $\gamma_{(t)}$ as
\begin{equation*}
\begin{aligned}
\bm{\theta}_{1_{(t+1)}} &=  \bm{\Sigma}_{u_{(t)}}^{-1} \bm{\mu}_{u_{(t)}} - \gamma_{(t)} \frac{\partial F}{\partial \bm{\psi}_{1_{(t)}}}, \\
\bm{\Theta}_{2_{(t+1)}} &= -\frac{1}{2}\bm{\Sigma}_{u_{(t)}}^{-1} - \gamma_{(t)} \frac{\partial F}{\partial \bm{\Psi}_{2_{(t)}}},
\end{aligned}
\end{equation*}
where $\partial F / \partial \bm{\psi}_{1_{(t)}} = \partial F / \partial \bm{\mu}_{u_{(t)}}$ and $\partial F / \partial \bm{\Psi}_{2_{(t)}} = \partial F / \partial \bm{\Sigma}_{u_{(t)}}$. The derivatives $\partial F / \partial \bm{\mu}_u$ and $\partial F / \partial \bm{\Sigma}_u$ are respectively expressed as \begin{equation*}
\begin{aligned}
\frac{\partial F}{\partial \bm{\mu}_{u}} =& \frac{1}{2} (\bm{\Omega}_{nu}^g)^{\mathsf{T}} \bm{\Lambda}_{nn}^{a'b} \bm{1} - (\bm{K}_{uu}^g)^{-1} (\bm{\mu}_{u} - \mu_0\bm{1}),  \\
\frac{\partial F}{\partial \bm{\Sigma}_{u}} =& -\frac{1}{4} (\bm{\Omega}_{nu}^g)^{\mathsf{T}} (\bm{\Lambda}_{nn}^{a'b} + \bm{I}) \bm{\Omega}_{nu}^g + \frac{1}{2} [\bm{\Sigma}_{u}^{-1} - (\bm{K}_{uu}^g)^{-1}], 
\end{aligned}
\end{equation*}
where $\bm{\Lambda}_{nn}^{a'b} = \bm{\Lambda}_{nn}^{a'} + \bm{\Lambda}_{nn}^b$, and $\bm{\Lambda}_{nn}^{a'}$ is a diagonal matrix with the diagonal element being
\begin{equation*}
[\bm{\Lambda}_{nn}^{a'}]_{ii} = [\bm{R}_g^{-1} (\bm{y} - \bm{\Omega}_{nm}^f \bm{\mu}_m) (\bm{y} - \bm{\Omega}_{nm}^f \bm{\mu}_m)^{\mathsf{T}} - \bm{I}]_{ii}.
\end{equation*}

For $q(\bm{f}_m) = \mathcal{N}(\bm{f}_m| \bm{\mu}_m, \bm{\Sigma}_m)$, the updates of $\bm{\mu}_{m_{(t+1)}}$ and $\bm{\Sigma}_{m_{(t+1)}}$ follow the foregoing steps, with the derivatives  $\partial F / \partial \bm{\mu}_m$ and $\partial F / \partial \bm{\Sigma}_m$ taking~\eqref{eq_partialF_mu_m_s_m}.

\section*{Acknowledgment}
This work is funded by the National Research Foundation, Singapore under its AI Singapore programme [Award No.: AISG-RP-2018-004], the Data Science and Artificial Intelligence Research Center (DSAIR) at Nanyang Technological University and supported under the Rolls-Royce@NTU Corporate Lab.

\ifCLASSOPTIONcaptionsoff
  \newpage
\fi



%

\bibliographystyle{IEEEtran}
\bibliography{IEEEabrv,DVSHGP}


%

\begin{IEEEbiography}[{\includegraphics[width=1in,height=1.25in,clip,keepaspectratio]{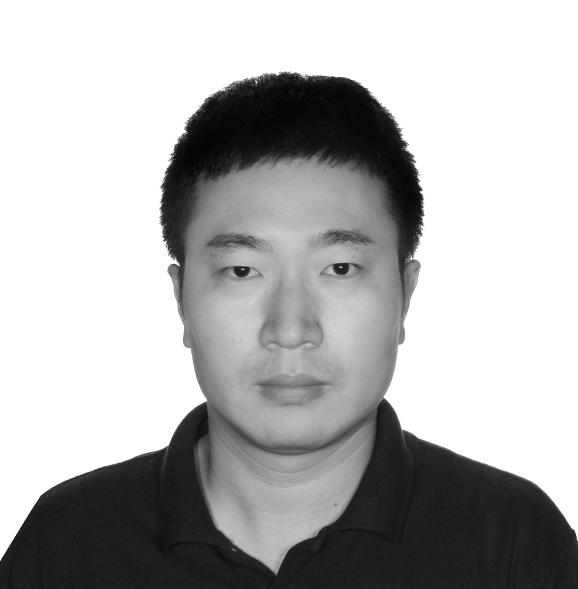}}]{Haitao Liu}received the Ph.D. degree from the School of Energy and Power Engineering, Dalian University of Technology, Dalian, China, in 2016. He is currently a Research Fellow with the Rolls-Royce@NTU Corp Laboratory, Nanyang Technological University, Singapore. His current research interests include multi-task learning, large-scale Gaussian process, active learning, and optimization.
\end{IEEEbiography}

\begin{IEEEbiography}[{\includegraphics[width=1in,height=1.25in,clip,keepaspectratio]{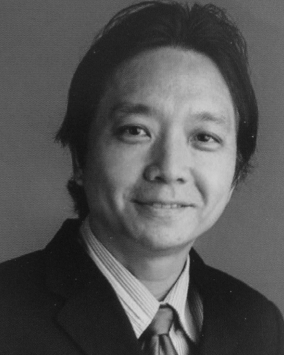}}]{Yew-Soon Ong} (M’99-SM’12-F’18) received the Ph.D. degree in artificial intelligence in complex design from the University of Southampton, U.K., in 2003. He is a President’s Chair Professor in Computer Science at the Nanyang Technological University (NTU), and holds the position of Chief Artificial Intelligence Scientist at the Agency for Science, Technology and Research Singapore. At NTU, he serves as Director of the Data Science and Artificial Intelligence Research Center and Director of the Singtel-NTU Cognitive \& Artificial Intelligence Joint Lab. His research interest is in artificial and computational intelligence. He is founding Editor-in-Chief of the IEEE Transactions on Emerging Topics in Computational Intelligence, Technical  Editor-in-Chief of Memetic Computing and Associate Editor of IEEE Transactions on Neural Networks \& Learning Systems, the IEEE Transactions on Cybernetics, IEEE Transactions on Evolutionary Computation and others. He has received several IEEE outstanding paper awards, listed as a Thomson Reuters highly cited researcher and among the World's Most Influential Scientific Minds. 
\end{IEEEbiography}


\begin{IEEEbiography}[{\includegraphics[width=1in,height=1.25in,clip,keepaspectratio]{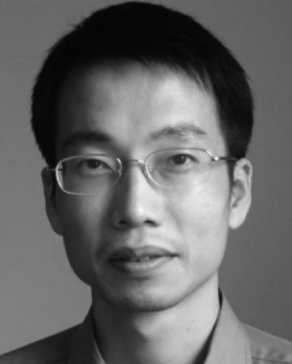}}]{Jianfei Cai} (S’98–M’02–SM’07) is a Professor at Faculty of IT, Monash University, where he currently serves as the Head for the Data Science \& AI Department. Before that, he was a full professor, a cluster deputy director of Data Science \& AI Research center (DSAIR), Head of Visual and Interactive Computing Division and Head of Computer Communications Division in Nanyang Technological University (NTU). His major research interests include computer vision, multimedia and deep learning. He has published more than 200 technical papers in international conferences and journals. He is a co-recipient of paper awards in ACCV, ICCM, IEEE ICIP and MMSP. He has served as an Associate Editor for IEEE T-IP, T-MM, T-CSVT and Visual Computer as well as serving as Area Chair for ICCV, ECCV, ACM Multimedia, ICME and ICIP. He was the Chair of IEEE CAS VSPC-TC during 2016-2018. He had also served as the leading TPC Chair for IEEE ICME 2012 and the best paper award committee co-chair for IEEE T-MM 2019.
\end{IEEEbiography}




\end{document}